\documentclass[10pt,journal,compsoc]{IEEEtran}
\newif\ifpeerreview
\peerreviewfalse

\usepackage[nocompress]{cite}
\usepackage{url}
\usepackage{amsmath,amssymb,graphicx}
\usepackage[switch]{lineno}
\usepackage{subcaption}

\DeclareMathOperator*{\fft}{FFT}
\DeclareMathOperator*{\ifft}{IFFT}

\newcommand{\paperID}{25}
\title{Super-Resolution with Structured Motion}

\author{Gabby~Litterio, Juan-David~Lizarazo-Ferro, Pedro~Felzenszwalb, Rashid~Zia
\IEEEcompsocitemizethanks{\IEEEcompsocthanksitem School of Engineering (Litterio, Felzenszwalb, Zia) and Department of Physics (Lizarazo-Ferro, Zia), Brown University, Providence, RI, 02906.  \protect\\
E-mail: gabby\_litterio@brown.edu, juan\_lizarazoferro@brown.edu, pedro\_felzenszwalb@brown.edu, rashid\_zia@brown.edu}}

\begin{document}

\IEEEtitleabstractindextext{%
\begin{abstract}
We consider the limits of super-resolution using imaging constraints. 
Due to various theoretical and practical limitations,
reconstruction-based methods have been largely restricted to small
increases in resolution.  In addition, motion-blur is usually seen as
a nuisance that impedes super-resolution.  We show that by using
high-precision motion information, sparse image priors, and convex
optimization, it is possible to increase resolution by large factors.
A key operation in super-resolution is deconvolution with a box.  In
general, convolution with a box is not invertible.  However, we obtain
perfect reconstructions of sparse signals using convex optimization.
We also show that motion blur can be helpful for super-resolution.  We
demonstrate that using pseudo-random motion it is possible to
reconstruct a high-resolution target using a single low-resolution
image.  We present numerical experiments with simulated data and
results with real data captured by a camera mounted on a computer-controlled stage.
\end{abstract}

}

\maketitle

\IEEEraisesectionheading{
  \section{Introduction}\label{sec:introduction}
}

\IEEEPARstart{W}{e} consider the super-resolution problem where the
goal is to reconstruct a high-resolution image from one or several
low-resolution images.  We focus primarily on the use of imaging
constraints and low-level image priors, and on the use of motion to
capture one or more pictures.

One of our motivations is to implement a compressed sensing system for
super-resolution.  Using a high-resolution CCD/CMOS sensor, it may be
possible to reconstruct an extremely high-resolution image using an
appropriate optical and computational system.  

Classical methods for super-resolution relate a collection of
low-resolution images to a high-resolution image using a model of the
imaging process.  Such methods have been largely restricted to relatively small
increases in resolution due to theoretical and practical limitations
\cite{baker_limits_2002, lin_fundamental_2004}.  In addition, motion
blur is usually seen as a nuisance that impedes super-resolution
\cite{ben-ezra_jitter_2004}.  

Despite the theoretical limits previously described in the literature, 
we show that it is possible to increase
resolution by large factors using sparse image priors.  We note a connection
between super-resolution and deconvolution with a box filter, which helps to
understand the ambiguities in the super-resolution problem.  In
particular, a collection of low-resolution images determines most of the Fourier
components of the super-resolution image and, in this setting, sparse
images can be recovered using convex optimization.

\begin{figure*}
    \centering
    \begin{tabular}{cccc}
    \boxed{\includegraphics[width=3.9cm]{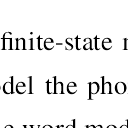}} &    
    \boxed{\includegraphics[width=3.9cm]{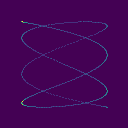}} &
    \boxed{\includegraphics[width=3.9cm]{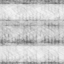}} &
    \boxed{\includegraphics[width=3.9cm]{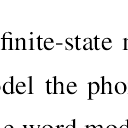}} \\
    Ground truth &   
    Sensor trajectory &
    Recorded image &
    Reconstruction \\
    $128 \times 128$ &
    $128 \times 128$ &
    $64 \times 64$ &
    $128 \times 128$ 
    \end{tabular}
    \caption{Using motion blur for super-resolution.}
    \label{fig:splash}
\end{figure*}

We also show that by using motion blur, it is possible to
reconstruct a high-resolution target from a \emph{single}
low-resolution image.  By moving the camera or sensor during an exposure,
we capture a superposition of multiple low-resolution images.  
Figure~\ref{fig:splash} illustrates the idea.  In this example where we capture an 
image while the camera/sensor undergoes a
large vibration.  We record a single low-resolution image and 
reconstruct a high-resolution target by solving a convex optimization problem.

Super-resolution imaging systems have many potential applications.  In
some settings, there are practical and physical limitations that lead
to low-resolution sensors.  Furthermore, there are many applications where a 
camera might be constantly moving, such as in aerial imaging.  
High-resolution imaging has
many applications in science and in archival/preservation work 
(including for art, documents, and biological samples).  For example,
\cite{ben-ezra_high_2010} describes a large format tile-scan camera
to image static scenes for cultural-heritage preservation.

For the case of visible light photography
with high-resolution sensors, our methods could be used to produce
gigapixel images using a small camera.   Pixel shift cameras are commercially
available and already use sensor motion to overcome the resolution limit
induced by color filters.  The same technology could be used to implement
the imaging methods described here.

We implemented a system to demonstrate some of our ideas using a
commonly available camera and computer-controlled motion stage.  The
experiments in Section~\ref{sec:experiments} demonstrate
results using both static images and images taken
with the camera moving at a constant velocity.

Throughout the paper, we focus on cases where the effective resolution
is limited by the sensor and not by the optical path (diffraction).
Even with high-resolution sensors, super-resolution can be useful for
obtaining a wider field of view with an appropriate lens. The
interlacing operation described in Section~\ref{sec:grid} is
reminiscent of a panorama, but interlaces a set of pictures taken with
small motion instead of tiling pictures taken with large motion, which
leads to a more compact design for a high-resolution camera.  


We assume throughout the paper that the sensor motion is known,
either because it is carefully planned and executed by a high-accuracy
actuator or because we have a high-accuracy readout of the sensor
position while it is being moved through some other means (such as
vibration).  In practice, the motion information should be accurate to the target pixel
resolution.   In Section~\ref{sec:scan}, we use a landmark (sharp edge) to
estimate a high-resolution camera position directly from a low-resolution image.

\section{Related Work}

The super-resolution problem has a long history (see,
e.g.\cite{irani-peleg-super,zisserman_capel_text,baker_limits_2002,lin_fundamental_2004}).
From a theoretical point of view, the results in
\cite{baker_limits_2002} and \cite{lin_fundamental_2004} are largely
negative, showing that ambiguity in the super-resolution problem grows
quickly with the desired increase in resolution.  In contrast, here we
note that with a particular sampling scheme, this ambiguity is
concentrated on relatively few Fourier coefficients.  Previous work in
compressed sensing \cite{candes_robust_2006,donoho1989-uncertainty}
has shown that sparse image priors can be used to resolve such
ambiguities.  

\cite{ben-ezra_jitter_2004} proposed the use of a "Jitter camera" to
improve the resolution of video by quickly moving the sensor in the
imaging plane of a camera without introducing motion blur, which was
believed to impede super-resolution.  Here we argue that motion blur
can in fact aid in super-resolution.

Intuitively, the combination of motion blur and discrete sampling can
be used to define a compressed sensing system.  The approach is
related to the use of random filters for compressed sensing
\cite{romberg2009-compressive,tropp2006random-filters}.

A variety of unique imaging system designs have been developed through
the use of compressed sensing methods.  Some physical implementations
include \cite{fergus2006-random-lens, antipa2017-diffusercam,
  duarte2008-single-pixel}.

Recently, some unusual imaging schemes have been proposed, not to achieve
super-resolution, but to remove motion-related artifacts.  For
example, parabolic lens motion \cite{levin_motion-invariant_2008},
induced by constant acceleration, generates a PSF that is invariant to
object velocity.  Additionally, fluttering a
camera shutter \cite{raskar2006-coded-flutter,tendero_flutter_2013}
has been shown to help in removing motion blur.  This leads to a
surprising idea that it may be beneficial to extend an exposure with a
coded shutter to compensate for motion blur.

Many super-resolution methods use training data or other sources of
information to improve image resolution.  The idea of exemplar based
super-resolution goes back to the work in \cite{baker_limits_2002} and
was implemented at a large scale in \cite{hayes-sun2012-super}.
Generic image priors based on self-similarity have also been used for
super-resolution \cite{glasner_super-resolution_2009}.  More recent
methods have used deep neural networks for super-resolution from a
single image\cite{dong2015-deep}.

In this paper, we focus on the use of total variation for
reconstruction.  In practice, alternative image priors, such as ones
based on self-similarity or deep networks, could also be implemented
using Plug-and-Play methods \cite{kamilov2023plug-play-priors}.  Total variation was
 also used as prior for super-resolution in \cite{zisserman_capel_text}.

Neural models have been increasingly used for image reconstruction,
including super-resolution \cite{xie_neural_2022}.  The use of a
neural radiance field (NeRF) allows for the estimation of a continuous
image instead of selecting a particular target resolution.  In this
case, one can implement the imaging model described here by Monte
Carlo integration via sampling a finite number of locations in the
imaging plane.

We consider only grayscale images to simplify the presentation and
experimental design.  For the case of color images captured with a
color filter array, one could reconstruct a super-resolution image for
each color channel separately, using measurements defined by pixels of
that color in the filter array.  This is related to the recent
approach in \cite{wronski2019-handheld} which performs
super-resolution without explicit demosaicing.  By using controlled
motion, one can also collect measurements for every pixel using each
color in the filter array.

\section{Imaging Model}

Our goal is to reconstruct a high-resolution image from one or several
low-resolution images.  In the mathematical model we consider here,
the low-resolution images are captured with a sensor that can
translate in the imaging plane of a camera.  In one setting, we
capture a series of images in a grid of high-resolution (sub-pixel)
locations.  In another setting, we capture one or more images while
the sensor is moving; for example, we can record a single or sequence
of pictures while vibrating the sensor or moving along a linear
trajectory (e.g, as may be the case in satellite imaging).

There are many equivalent practical situations.  In the experiments in
Section~\ref{sec:experiments}, we use a standard camera on a moving
stage, which is equivalent to moving the sensor when viewing a planar
object parallel to the camera.  In a microscope, instead of moving the
sensor, we can precisely move the sample using a piezo nanopositioning
stage.

Let $g : \mathbb{R}^2 \rightarrow \mathbb{R}$ denote a continuous
brightness function in the imaging plane of a camera.  We do not model
the optics of the camera and focus only on measuring the image
projected onto the imaging plane.

An $n \times m$ sensor integrates $g$ over square pixels tiling a
rectangular area.  We use $\Delta$ to denote the side length of a
pixel in the sensor.  Each pixel integrates the light that falls in a
particular square region over a unit of time.

Ignoring measurement noise, if we place the sensor in the imaging
plane with bottom-left corner at $(x_0,y_0)$ we capture a discrete
image $I$ with
$$I[i,j] = \int_0^\Delta \int_0^\Delta g(x_0+i\Delta+x,y_0+j\Delta+y)
dx dy.$$

We will also consider the case where the sensor moves while the image
is captured.  If the sensor follows a trajectory $p(t) = (x_0(t),y_0(t))$, we
capture an image $I$ with
$$I[i,j] = \int_0^1 \int_0^\Delta \int_0^\Delta g(x_0(t)+i\Delta+x,y_0(t)+j\Delta+y)
dx dy dt.$$

Note that translating the sensor by $z$ is equivalent to translating
$g$ by $-z$.  Let $q(x,y)$ denote the occupancy map defined by the
trajectory $-p(t)$.  That is, $q(x,y)$ is the fraction of time
the trajectory spends at position $(x,y)$.  The image $I$ captured by the moving
sensor is equivalent to the image of $g \otimes q$ captured by a static
sensor.

We consider methods that increase the resolution of the camera sensor
to generate a super-resolution image $J$ that would have been recorded
by a static sensor with pixels of size $\Delta/f$ for an integer
factor $f$ defining the resolution increase.  This virtual sensor
covers an area that is at least as big as the camera sensor and the
image $J$ has at least $fn \times fm$ pixels (the area may be larger
due to the sensor motion).  In our experiments, we have used
super-resolution factors as high as $f=8$.

\section{Grid of Images}

Let $J$ be an image generated by a (virtual) sensor with pixel size
$\Delta/f$ at the origin.  Let $I$ be an image captured by a
(physical) sensor with pixel size $\Delta$ at $(x_0,y_0)$ where
$$x_0 = k (\Delta/f),$$
$$y_0 = l (\Delta/f),$$ with $k$ and $l$ integers.  In this case, the
low-resolution pixels in $I$ are aligned with the high-resolution
pixels in $J$.  Therefore, each value in $I$ is the sum of $f^2$
values in $J$,
$$I[i,j] = \sum_{i'= 0}^{f-1} \sum_{j'=0}^{f-1} J[k+fi+i',l+fj+j'].$$
We see that $I$ is a decimation (subsampling) of $J \otimes B$ where
$B$ is a two-dimensional box filter of width $f$.

Now suppose we capture $f^2$ images $\{I_{k,l}\}$
by translating the sensor over an $f \times f$ grid of high-resolution
locations,
$$L = \{(k(\Delta/f), l(\Delta/f)) \,|\, 0 \le k \le f-1, 0 \le l \le
f-1 \}.$$ Capturing such images involves moving the sensor to
sub-pixel locations defined by a step size $\Delta/f$.  The total
vertical and horizontal displacement required to capture all images is
less than $\Delta$.

\subsection{Interlacing} 

The images $\{I_{k,l}\}$ can be rearranged into a
single $fn \times fm$ image $H$ such that $H = J \otimes B$, where $B$
is a box of width $f$.  Since each image $I_{k,l}$ is a decimation of
$J \otimes B$, the rearrangement interlaces the captured images:
$$H[k+if,l+jf] = I_{k,l}[i,j].$$

\begin{figure*}
\includegraphics[width=\textwidth]{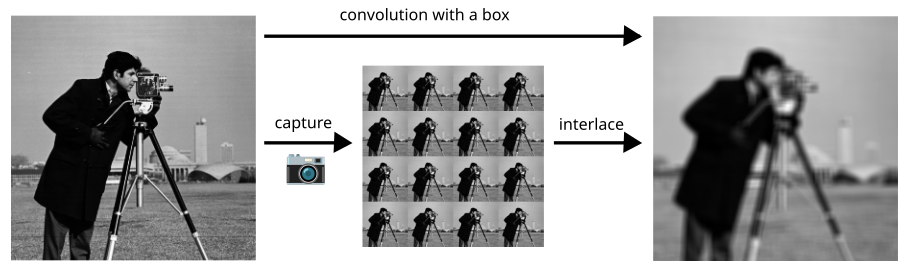}
\caption{Interlacing $f \times f$ low-resolution images taken in a
  grid of sub-pixel locations, we obtain the convolution of the
  high-resolution target with a two-dimensional box filter of width
  $f$.}
\label{fig:interlacing}
\end{figure*}

Figure~\ref{fig:interlacing} illustrates the process where we capture
multiple low-resolution images at sub-pixel shifts 
and interlace the result to obtain a single image $H$.  
The interlaced image $H$ has $fn \times
fm$ pixels and defines the imaging constraints on $J$ through the
relation $H = J \otimes B$.

\subsection{Deconvolution with a Box}

Recovering the super-resolution image $J$ is equivalent to a
deconvolution of $H$ with a two-dimensional box filter of width $f$.
That is, we need to de-blur $H$ to recover $J$.

\begin{figure*}
\centering
\begin{subfigure}{0.4\textwidth}
\includegraphics[width=2.5in]{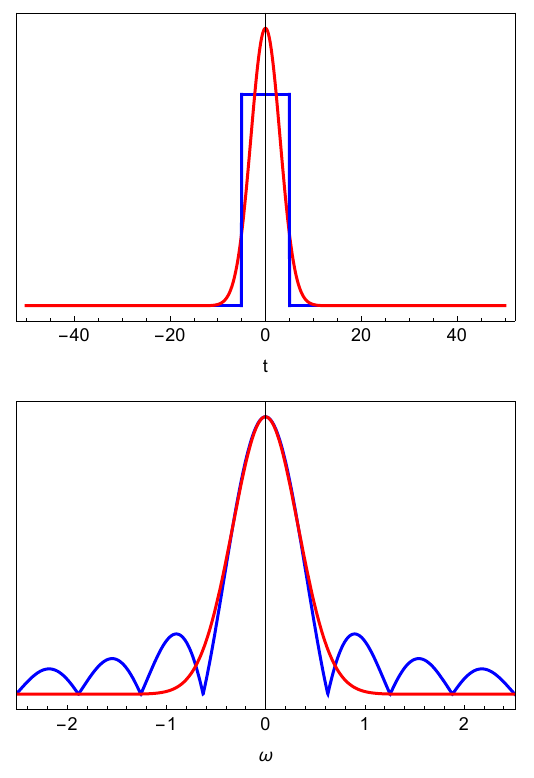} 
\caption{}
\end{subfigure}
\begin{subfigure}{0.4\textwidth}
\includegraphics[width=3.5in]{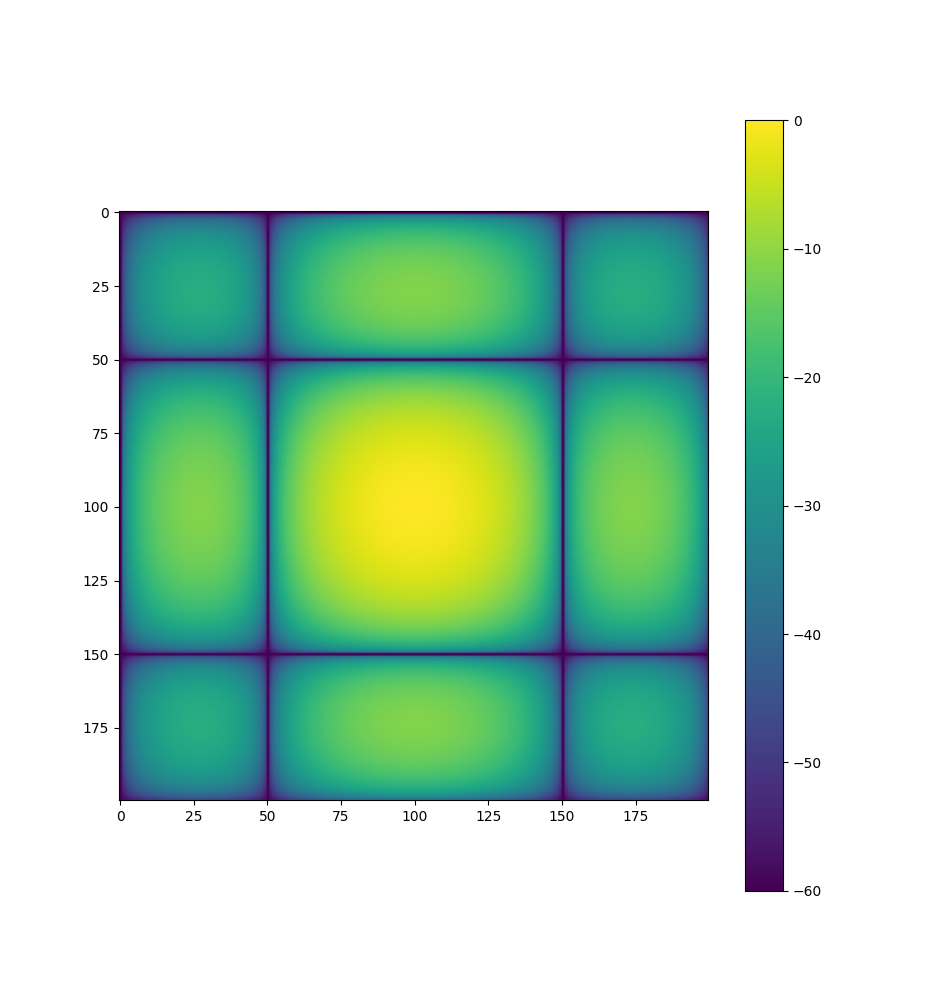} 
\caption{}
\end{subfigure}
\caption{(a) One-dimensional box and Gaussian (top) and their
  Fourier transforms (bottom). The red plots correspond to a Gaussian
  and the blue plots correspond to a box.  (b) Bode plot (magnitude of
  Fourier transform in dB) of a two-dimensional discrete box filter of width
  4.}
\label{fig:sinc-vs-gaussian}
\end{figure*}

Note that a box filter is not a good smoothing or low-pass filter.
Convolution with a box yields a signal that is visually blurry but
still has quite a lot of high-frequency information.  In
Figure~\ref{fig:sinc-vs-gaussian}(a) we compare the Fourier
transform of a one-dimensional box and a Gaussian of similar width.
The Fourier transform of a box is a sinc, and the Fourier transform of a Gaussian is a Gaussian.  Although the sinc has isolated
zeros, the magnitude of the Fourier transform of the box decays
much slower than the magnitude of the Fourier transform of 
a Gaussian.

For a two-dimensional discrete (and periodic) signal, we
illustrate the magnitude of the DFT of a box of width $4$ in
Figure~\ref{fig:sinc-vs-gaussian}(b). Note that there are relatively
few coefficients in the DFT of $B$ with magnitude close to 0.  We can
therefore directly recover most of the coefficients in the Fourier
transform of $J$ from the Fourier transform of $H$, except for
isolated frequencies where the Fourier transform of $B$ is zero or
close to zero.

This is exactly
the setting in which sparse image priors and convex optimization are known to
yield surprisingly good (``perfect'') reconstructions
\cite{candes_robust_2006,donoho1989-uncertainty}.

For one-dimensional sparse signals, 
deconvolution with a box using $\ell_1$
regularization leads to perfect reconstructions \cite{felzenszwalb_2024}.  This is true \emph{despite} the ambiguity
in the imaging constraints due to the unknown Fourier coefficients.  
The key here is the choice of prior which
selects a particular solution among all solutions that agree (or nearly agrees in the case of noisy data) with the measurements.

\begin{figure*}
\centering
\begin{subfigure}{0.32\textwidth}
\centering
\includegraphics[trim={1.2cm, 1.8cm, 1.2cm, 0.6cm}, clip, height=2in]{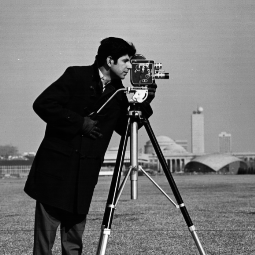} 
\caption{Input ($J$)}
\end{subfigure}
\begin{subfigure}{0.32\textwidth}
\centering
\includegraphics[trim={0.15cm, 0.225cm, 0.15cm, 0.075cm}, clip, height=2in]{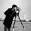} 
\caption{1 of 64 low-res images ($I_{k,l}$)}
\end{subfigure}
\begin{subfigure}{0.32\textwidth}
\centering
\includegraphics[trim={1.2cm, 1.8cm, 1.2cm, 0.6cm}, clip,height=2in]{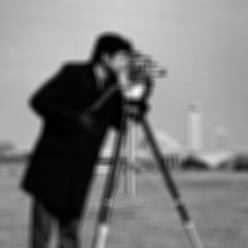} 
\caption{Interlaced images ($H$)}
\end{subfigure}
\vspace{.1cm}

\begin{subfigure}{0.32\textwidth}
\centering
\includegraphics[trim={1.2cm, 1.8cm, 1.2cm, 0.6cm}, clip,height=2in]{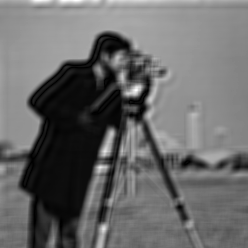} 
\caption{Wiener deconvolution}
\end{subfigure}
\begin{subfigure}{0.32\textwidth}
\centering
\includegraphics[trim={1.2cm, 1.8cm, 1.2cm, 0.6cm}, clip,height=2in]{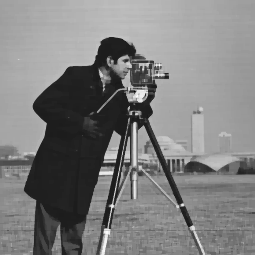} 
\caption{TV deconvolution}
\end{subfigure}
\caption{Super-resolution with $f=8$.  By interlacing 64 low-resolution images, we obtain the measurements $H = J \otimes B$.  
 Deconvolving $H$ with a TV prior leads to an almost perfect reconstruction.}
\label{fig:MIT}
\end{figure*}

Figure~\ref{fig:MIT} illustrates a numerical experiment with a
simulated camera.  The deconvolution of $H$ using a total variation
prior yields an almost perfect reconstruction including very
high-resolution details.

\subsection{The Limits of Super-Resolution}
\label{sec:limit}

Previous work \cite{baker_limits_2002} considered the setting discussed here 
and showed that the ambiguity in the super-resolution
problem grows quickly with the super-resolution factor $f$.  
This was characterized in terms of the volume of the set of solutions that are
compatible with a set of images and 
the condition number of a resulting linear system.

We note, however, that the negative results in \cite{baker_limits_2002} do
not account for the use of a prior to select a particular solution among
the ones that are compatible (or nearly compatible) with the measured data.  
For the experiments with imaging constraints 
in \cite{baker_limits_2002} the ambiguity was resolved with a quadratic 
smoothness prior, while we use total variation.  

The difference between using a quadratic smoothness prior and total variation
may seem subtle at first degree, but one of the most striking results in compressed sensing 
is that priors that induce sparsity, like total variation, can lead to perfect reconstructions with under-determined linear systems \cite{candes_robust_2006}.  The fact that natural images
have the appropriate sparsity structure is a surprising phenomenon of its own.  Total variation is also
known to preserve/recover sharp edges in image restoration (see, e.g. \cite{chambolle2010introduction}).

\begin{figure*}
\centering
\begin{subfigure}{0.32\textwidth}
\centering
\includegraphics[height=2.2in]{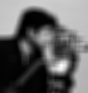} 
\caption{Interlaced images ($H$)}
\end{subfigure}
\begin{subfigure}{0.32\textwidth}
\centering
\includegraphics[height=2.2in]{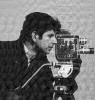} 
\caption{Quadratic prior}
\end{subfigure}
\begin{subfigure}{0.32\textwidth}
\centering
\includegraphics[height=2.2in]{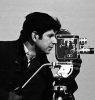} 
\caption{TV prior}
\end{subfigure}
\caption{Deconvolution of $H$ with a quadratic smoothness prior leads to an overly smooth image and ringing artifacts while using a TV prior yields an almost perfect reconstruction.}
\label{fig:limit}
\end{figure*}

Figure~\ref{fig:limit} compares a result that minimizes a quadratic smoothness 
prior (as in \cite{baker_limits_2002}) to a result that minimizes the total variation.  
In this case there is no measurement error and both solutions satisfy the imaging constraints 
defined by $H = J \otimes B$.  This illustrates how the
solution is ambiguous (as expected) and how the
choice of image prior can have a significant impact on the quality of the 
results.   Moreover, the example
illustrates that a low-level prior is sufficient to resolve high-resolution details.

\subsection{Wiener Deconvolution}

Using a Gaussian prior for the Fourier coefficients of the 
high-resolution image leads to a very fast Wiener deconvolution 
algorithm via the FFT and an inverse transform.  

The Wiener filter computes a MAP estimate of the 
super-resolution image $J$.  
The algorithm is defined by a single parameter $\gamma$ used to 
shrink the estimated Fourier coefficients.

\begin{align*}
\bar{B} & = \fft(B) \\
\bar{H} & = \fft(H) \\
\bar{J} & = \frac{\bar{H}}{\bar{B} + \gamma} \\
J & = \ifft(\bar{J})
\end{align*}

\subsection{Sparse Reconstruction with TV prior}

As discussed above, previous work has shown that sparse signals can
often be perfectly or nearly perfectly recovered from 
incomplete frequency information \cite{candes_robust_2006,donoho1989-uncertainty}.

In practice, we regularize the reconstruction of the super-resolution
image $J$ using the total variation (TV) measure.  This leads to a
convex optimization problem,
$$\min_J ||H - (J \otimes B)||_2^2 + \lambda TV(J)$$
where $TV(J)$ is the total-variation of $J$,
$$TV(J) = \sum_{i,j} |J[i,j]-J[i+1,j]| + |J[i,j]-J[i,j+1]|.$$
and $\lambda$ is a regularization parameter.

\section{Super-resolution by Deblurring}

In the previous section, we considered the case of multiple static images taken from different sensor locations.
Here we consider the case where one or more images are captured while the sensor is moving.  

\begin{figure}
    \centering
    \includegraphics[width=0.99\linewidth]{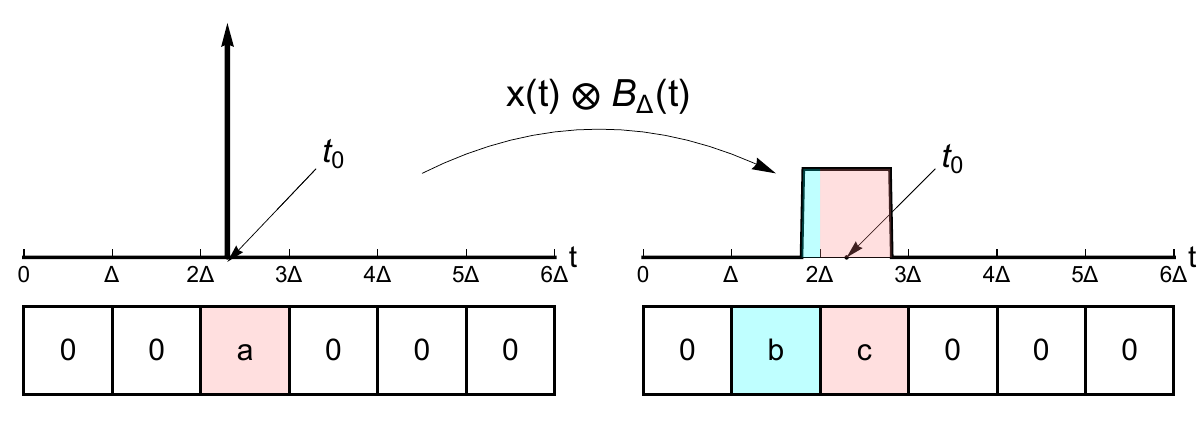}
    \caption{We can localize a point source to sub-pixel location if
      the camera undergoes constant velocity 
      motion of one pixel during capture.  The motion spreads the source to two
      pixels, and the
      relative values determine the source location.  On the left, we
      have the continuous brightness function and the image captured
      with a static sensor.  On the right, the sensor moves with
      constant velocity by one pixel during capture.  The effect of the
      sensor motion is equivalent to a convolution of the brightness 
      function with a box.}
    \label{fig:point-source}
\end{figure}

First, we note that a small amount of motion blur can be helpful for
super-resolution.  Consider the problem of localizing a point source
in one dimension; without motion, we can only localize the point
source to the pixel size.  Now suppose we move the sensor/camera with
constant velocity by exactly one pixel while taking a picture.
Figure~\ref{fig:point-source} illustrates this situation.  The motion leads to a convolution of the brightness function with a box.  The point
source is spread over two pixels and we observe two non-zero values
$b=I[k]$ and $c=I[k+1]$.  Together, these measurements determine both the
total light intensity and the sub-pixel point source location,
$$a=b+c,$$
$$t_0 = (k+0.5)\Delta + c/(b+c)).$$ Optical blurring, such as due to
diffraction, can lead to a similar effect (see,
e.g.\cite{schiebinger2018superresolution-separation}) but the use of
motion may have advantages since, as discussed in the previous
section, convolution with a box preserves more information when
compared to a Gaussian filter of similar width.

\begin{figure*}
\centering
    \begin{subfigure}{0.19\textwidth}
    \boxed{\includegraphics[width=3.3cm]{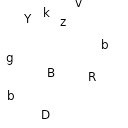}}
    \end{subfigure}
    \hfill
    \begin{subfigure}{0.19\textwidth}
    \boxed{\includegraphics[width=3.3cm]{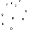}} 
    \end{subfigure}
    \hfill
    \begin{subfigure}{0.19\textwidth}
    \boxed{\includegraphics[width=3.3cm]{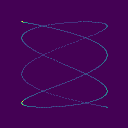}} 
    \end{subfigure}
    \hfill
    \begin{subfigure}{0.19\textwidth}
    \boxed{\includegraphics[width=3.3cm]{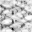}}
    \end{subfigure}
    \hfill
    \begin{subfigure}{0.19\textwidth}
    \boxed{\includegraphics[width=3.3cm]{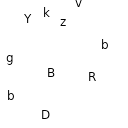}} 
    \end{subfigure}
\vspace{.1cm}

    \begin{subfigure}{0.19\textwidth}
    \boxed{\includegraphics[width=3.3cm]{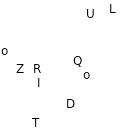}}
    \caption{$J$ ($128 \times 128$)}
    \end{subfigure}
    \hfill
    \begin{subfigure}{0.19\textwidth}
    \boxed{\includegraphics[width=3.3cm]{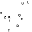}}
    \caption{static $I$ ($32 \times 32$)}
    \end{subfigure}
    \hfill
    \begin{subfigure}{0.19\textwidth}
    \boxed{\includegraphics[width=3.3cm]{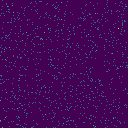}}
    \caption{trajectory}
    \end{subfigure}
    \hfill
    \begin{subfigure}{0.19\textwidth}
    \boxed{\includegraphics[width=3.3cm]{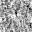}}
    \caption{moving $I$ ($32 \times 32$)}
    \end{subfigure}
    \hfill    
    \begin{subfigure}{0.19\textwidth}
    \boxed{\includegraphics[width=3.3cm]{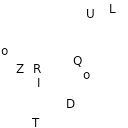}}
    \caption{Reconstructed $J$}
    \end{subfigure}
    \hfill
    
    \caption{(a) High-resolution targets with 10 random characters. (b) Images recorded by a static low-resolution sensor.  (c)
      Sensor trajectory (vibration on the top and random
      translations on the bottom).  (d) Low-resolution images
      recorded by a moving sensor.  (e)
      Reconstruction of the high-resolution target from the image recorded by the moving sensor.}
    \label{fig:font-simulation}
\end{figure*}

We also note that for ``sparse'' scenes, we can use \emph{large}
motions to (in essence) take multiple low-resolution images in
different areas of the sensor.  
Figure~\ref{fig:font-simulation} illustrates a simulation where the sensor
undergoes a large magnitude vibration (top row) while capturing an image.  
This leads to the somewhat
counterintuitive notion that shaking the camera during an exposure can
lead to sharper pictures.

Being able to take pictures while the sensor/camera is moving allows
for faster acquisitions, which is often important.  In this case,
there is no need to wait for the sensor to move and stabilize in one
position before taking a picture; instead, we can follow a continuous
trajectory and take a sequence of pictures along the way.

As in the previous section, let $J$ be a super-resolution image
generated by a virtual sensor at the origin.  Now let $I$ be an
image captured by a moving sensor.  Let $p(t)=(k(t),l(t))$ for $0 \le t \le
1$ denote a sensor trajectory on the grid of
high-resolution pixel locations.  

Let $B$ be a two-dimensional box of width $f$ and $Q$ be the occupancy map of the 
trajectory $-p(t)$.  
When the sensor undergoes continuous motion we use linear
interpolation to define a discrete high-resolution occupancy map $Q$.  
  The relationship between $J$ and $I$ is captured by
a decimation by factor $f$ of the sequential convolution of $J$ with
$Q$ and $B$.  The convolution of $J$ with $Q$ accounts for the sensor
motion, while the convolution with $B$ and decimation accounts for the
relative size of the physical and virtual pixels.

Let $(J)\downarrow_f$ denote the decimation of $J$ by a factor $f$.
The imaging constraints are summarized by the linear system,

$$I = (J \otimes Q \otimes B)\downarrow_f.$$

Due to the decimation by $f$, this system of equations has many fewer constraints than
variables.  The convolution with $B$ also introduces ambiguities.  Nonetheless,
these constraints can lead to near perfect reconstructions with an appropriate choice of
image prior as we demonstrate in simulations below.

In some applications we may have a sequence of low-resolution images
$I_1,\ldots,I_s$ taken over one or multiple trajectories.  Let $Q_j$
denote the occupancy map of the trajectory followed during the
acquisition of $I_j$.

We can reconstruct a high-resolution image using the data from
multiple motion blurred images simultaneously.  To estimate a sparse
image using $\ell_1$ regularization we solve the convex optimization problem,
$$\min_J \sum_{k=1}^s || I_k - (J \otimes Q_k \otimes B)\downarrow_f ||_2^2 + \lambda ||J||_1.$$
Similarly we can regularize the reconstruction using TV,
$$\min_J \sum_{k=1}^s || I_k - (J \otimes Q_k \otimes B)\downarrow_f ||_2^2 + \lambda TV(J).$$
Here again $\lambda$ is a regularization parameter.

\subsection{Simulations}

To demonstrate the use of motion blur for super-resolution, we simulated
images obtained using different sensor trajectories.  
For the experiments in this section, we treat the high-resolution
target as periodic. The sensor motion therefore leads to a cyclic
convolution between the target and the sensor occupancy map.  We used
the $\ell_1$ norm of the target image $J$ to regularize the
reconstructions.

Figure~\ref{fig:font-simulation}(a) shows two examples of the
high-resolution targets used in the experiments.  Each target is a
$128 \times 128$ image with several random characters placed in random
locations.  Figure~\ref{fig:font-simulation}(b) shows the results of
taking a low-resolution picture of each target using a $32 \times 32$
sensor.  In this case, there is no sensor motion.  We can see that
with such a low-resolution picture, the characters become completely
unreadable.

Figure~\ref{fig:font-simulation}(d) shows the results of imaging the
two targets from Figure~\ref{fig:font-simulation}(a) with a moving
sensor.  In each case, the sensor undergoes a large
motion. Figure~\ref{fig:font-simulation}(e) shows the reconstructions
of the targets from the low-resolution image captured while the
sensor is moving.  In these examples, we obtain a nearly perfect
high-resolution image from a single low-resolution image by
leveraging motion blur.

We consider the following trajectories for the sensor:

\begin{enumerate}
    \item Vibration (vibration).  The $x$ and $y$ coordinates of the
      moving sensor are defined by two sinusoidals with large magnitude.  (Figure~\ref{fig:font-simulation} top row)

    \item 1000 random displacements (shifts).  In this case, the
      trajectory is a sequence of 1000 random locations.  Each
      location is selected independently from a uniform distribution
      in the high-resolution grid.
    (Figure~\ref{fig:font-simulation} bottom row)
      
    \item Random walk with 1000 steps (walk).  In this case, the
      sensor undergoes a random walk in the grid of high-resolution
      locations.  (Not shown)
\end{enumerate}

\begin{figure}
    \centering
    \includegraphics[width=0.50\textwidth]{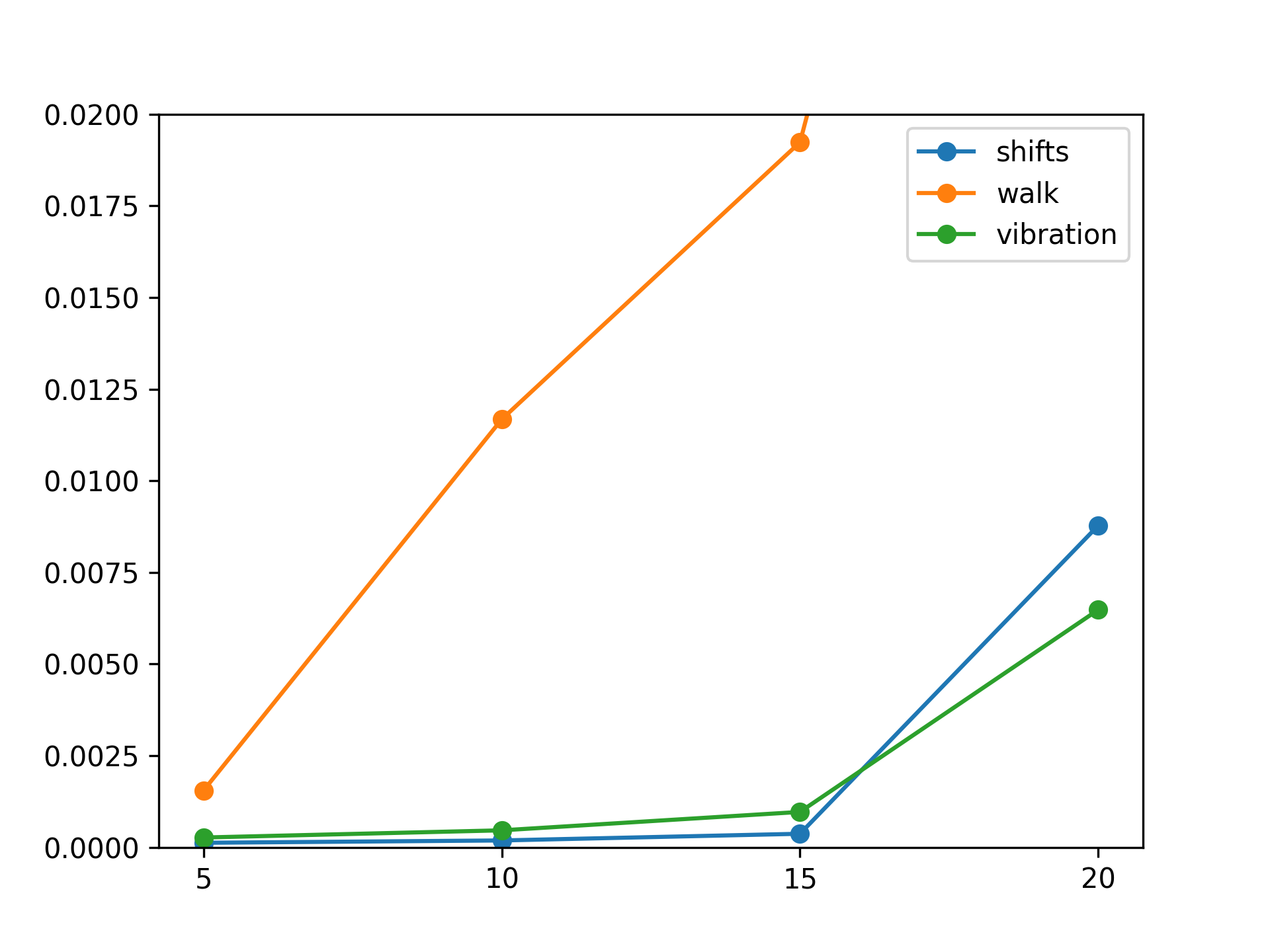}
    \caption{Mean reconstruction error for targets with different
      numbers of characters.  Each target is 128x128 and we capture a
      single 32x32 image undergoing motion.}
    \label{fig:font-rms}
\end{figure}

Figure~\ref{fig:font-rms} shows the mean RMS error obtained with each
trajectory for targets of increasing complexity.  We see that for
sparse targets, where the number of characters in the image is small,
it is possible to obtain perfect 128x128 images using a single 32x32
blurry picture.

\section{Experimental Results}
\label{sec:experiments}

We implemented the reconstruction algorithms in python.  We used numpy for
the Wiener deconvolution method and both cvxpy and pylops for
computing reconstructions using sparse image priors.  We found that using
the OSQP solver within cvxpy yields good results with small images, as
seen in the simulations throughout the paper.  For the experiments
with real data, we used the split-bregman algorithm implemented in
pylops due to the computational demands of processing larger images.

All of the experiments were done with a desktop computer with a 13th
Gen Intel(R) Core(TM) i5-13400F CPU with 16 cores and 32GB of RAM.

When $f=8$, reconstructing a 1200x1520 image from 64 separate 150x190
low-resolution images with the Wiener deconvolution algorithm took
about 0.4 seconds.  Computing a TV reconstruction using pylops with
the same data took between 5 and 10 minutes depending on various
tolerances and algorithm parameters.  

All of the running times scale linearly with
the image sizes.  A direct reconstruction of a very large (gigapixel) image 
using a sparse image prior would be impractical with our current implementation.
However, we could break the image into (overlapping) tiles and reconstruct each tile independently in parallel.

\subsection{Setup}

The data used in the following reconstructions consist of sets of
images captured by a monochrome 1.6 megapixel (1440x1080) CMOS camera
(Thorlabs Zelux) with 3.45 $\mu$m square pixels and a wide-angle 8 mm
f1.8 Edmund Optics lens opened to maximum aperture and focused at
approximately 3 meters.  This particular choice of lens and sensor
leads to an optical system that is not diffraction limited despite the
relatively small pixel size.

The pixels in our camera are smaller than many professional 
photography cameras, which are also not diffraction limited when using large 
apertures.  Pixel shift technology that is available on various
cameras could in principle be used to implement the imaging systems described 
here, although that would likely require special access to the control system 
of the camera.   

For the experiments described here, instead of moving the sensor, we move
the camera.  The camera was mounted on a movable stage and controlled by a nanopositioning system (Nanomotion II) that uses stepper motors to move the stage at nanometer level resolution in two separately controlled axes. Movements made by the stage have high precision and repeatability, though no position readout is available. The stage is connected to and controlled by a computer via a serial port. 

A static platform was mounted 3 meters away and parallel to the camera
to hold targets to be imaged. The illumination source was a battery
powered LED tube light (10 inch PavoTubeII 6C RGBWW).

\begin{figure*}
\centering
\label{fig:Stage and Camera Setup}
\begin{subfigure}{0.3\textwidth}
\includegraphics[width=2.13in]{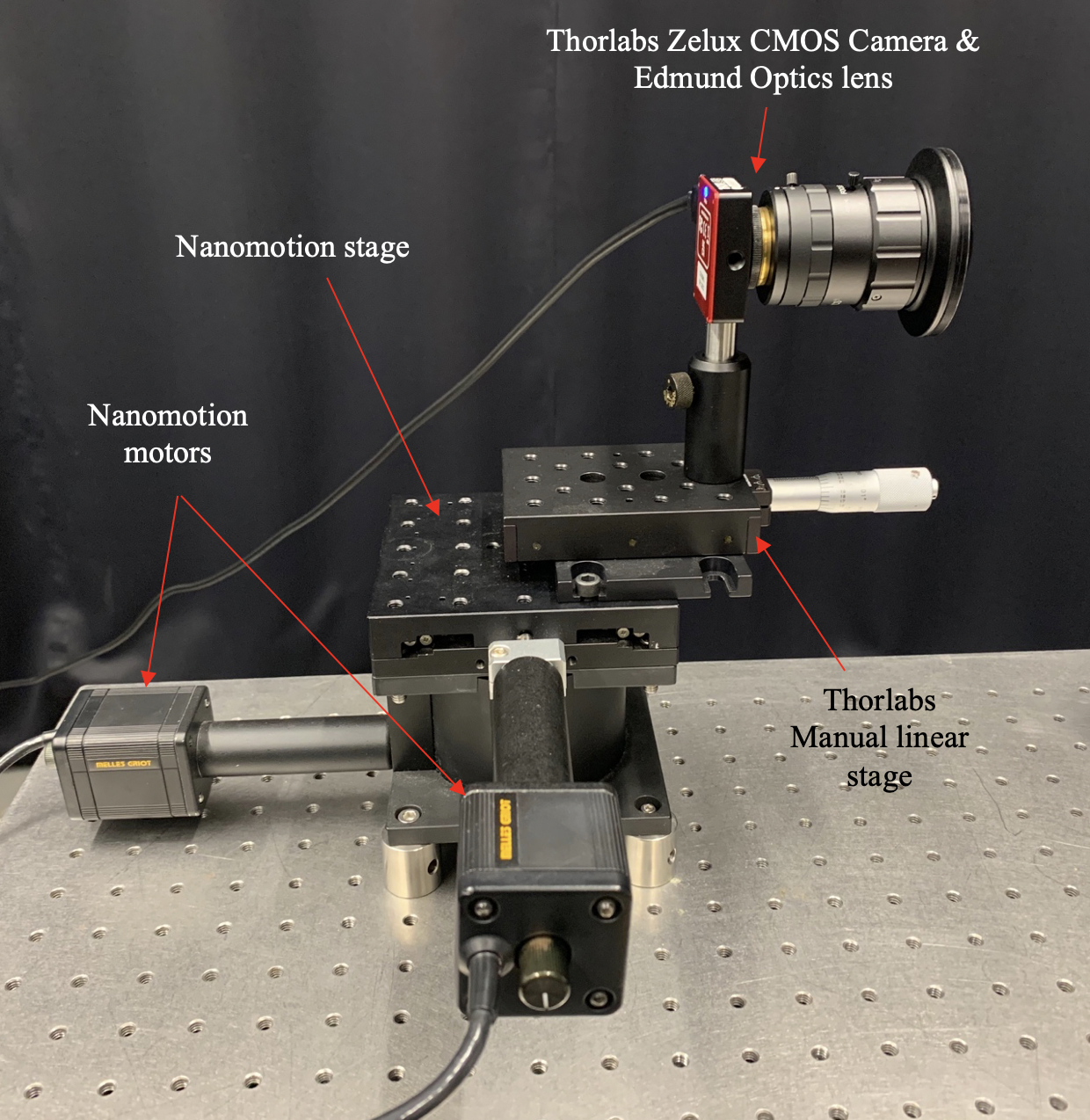}
\caption{Camera and stage setup}
\end{subfigure}
\begin{subfigure}{0.65\textwidth}
\includegraphics[width=4.5in]{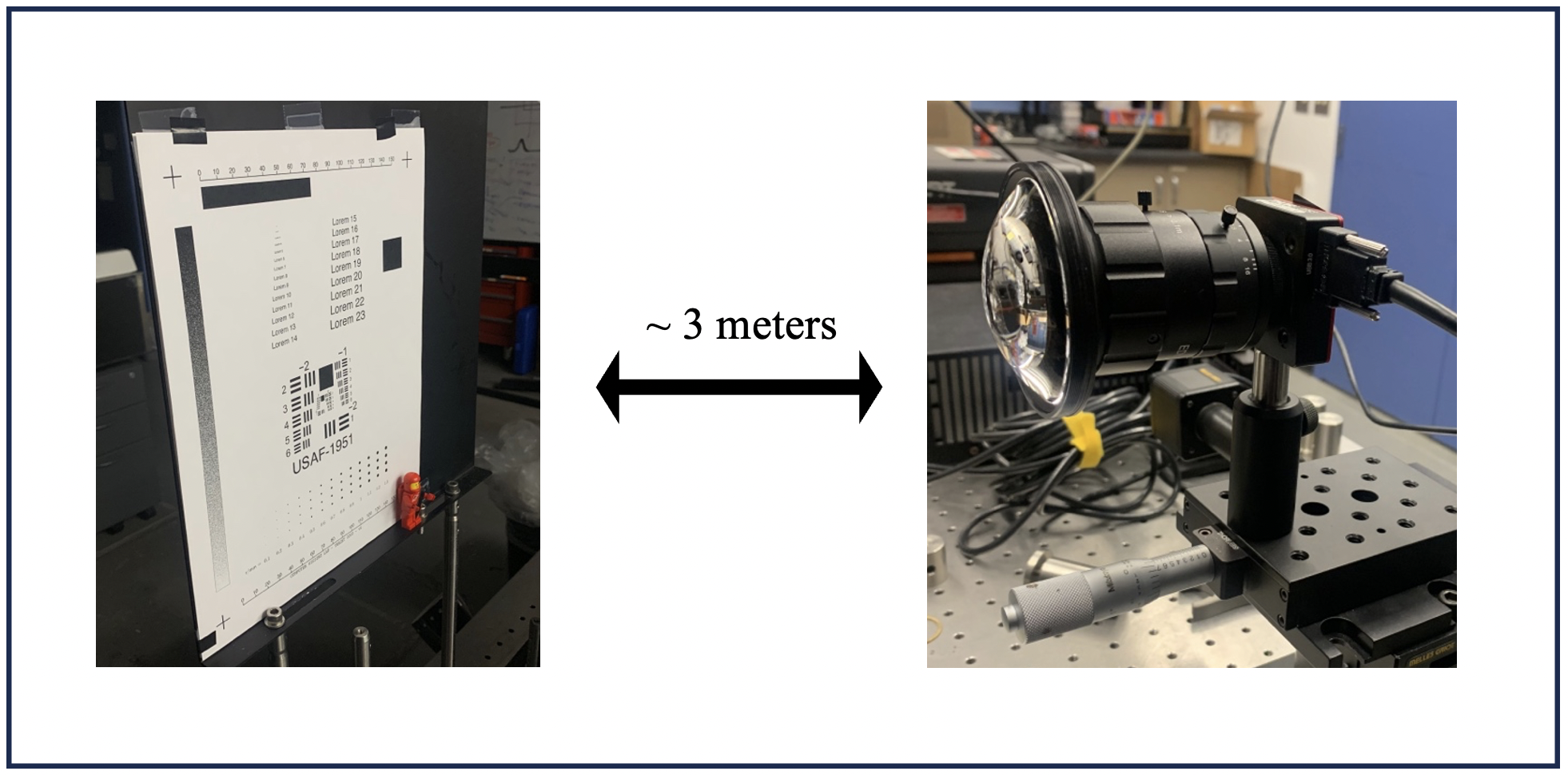}
\caption{Camera is aimed at a target platform }
\end{subfigure}
\caption{(a) The camera is attached to a linear stage that adjusts the
  distance between the camera and the target. The linear stage is
  secured to the Nanomotion II stage. (b) Example of printed design
  and small object placed on the target platform.}
\end{figure*}

We imaged a variety of targets including small three-dimensional
objects (ex. Ketchup packet and LEGO(R) sets) and multiple 8.5'' x
11'' pages of paper containing designs printed by a standard black and
white laser printer, such as a page of text and a to-scale version of
the 1951 USAF resolution target.

The USAF target is a standard test chart used to determine the
resolution of an optical system. It consists of vertical and
horizontal sets of three bars at varying sizes and spatial frequencies
that are organized in numbered groups, each with six labeled
sets. Once imaged, intensity profiles across the bars can be analyzed;
past the resolution limit, three distinct peaks/valleys will no longer
be visible in the profile. The group and element number of the
smallest resolvable set is used to look up the width and spacing of
these bars, and this value, in micrometers or lp/mm, is a measure of
the resolution of the system.

For each set of experiments, flat-field images were acquired by
imaging a blank sheet of paper, which served as the background for all
targets.

Our system is highly controlled via a stationary target,
fixed lighting, and long acquisition times.  One application where these
conditions are available is in digital archiving.  In this field, unique camera architectures have been developed to capture high-resolution images (see, e.g. \cite{ben-ezra_high_2010}).   

\subsection{Grid Exposures}
\label{sec:grid}

To acquire a static grid of images, the stage was moved through a series of
positions in a plane parallel to the target, coming to a stop at each
position and waiting for two frames to be captured, before
continuing. The overall grid motion was repeated five times. Averaging
many frames reduces the effects of sensor noise and running multiple
trials of the grid sequence minimizes the effects of temporal
illumination variation during the acquisition.

The number of acquired low-resolution images is determined by the
desired resolution increase factor $f$.  The actual position of these
acquired images depends on the magnification factor induced by the
camera optics and the distance to the target.

Moving the camera is equivalent to moving the target, and the
magnification $m$ of the system relates the distance traveled by the
camera to the distance traveled by the image on the sensor.  We can
determine $m$ using the optical properties of the camera and the
distance to the target.  In practice, we find $m$ by imaging an object
with known length and calculating its size on the sensor by counting
the number of pixels it spans and multiplying by the pixel size.

With grid capture (see Section~\ref{sec:grid}), the image of the
target moves by $\frac{\Delta}{f}$ between adjacent grid locations, for a
total of $f \times f$ locations.  Taking into account the
magnification $m$, the distance the stage travels between neighboring
grid positions, i.e. step size, is $\frac{\Delta}{fm}$.

For $f=8$, we use 64 grid locations. The magnification of our system
was calculated to be 0.002484 (we use a wide-angle lens), which gives
a step size of 0.1736 mm.

Due to the large field of view of the camera, our targets occupy a
relatively small (150x190) area in the images acquired, which means
that, for the experiments presented here, we are only using a small
region of the physical sensor in the camera.  In other applications,
the same setup would allow for capturing very wide field of views.

\begin{figure*}
\centering
\begin{subfigure}{0.32\textwidth}
\includegraphics[trim={1cm, 1.2cm, 1cm, 2cm}, clip, width=2.2in]{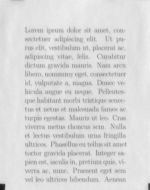} 
\end{subfigure}
\begin{subfigure}{0.32\textwidth}
\includegraphics[trim={8cm, 9.6cm, 8cm, 16cm}, clip, width=2.2in]{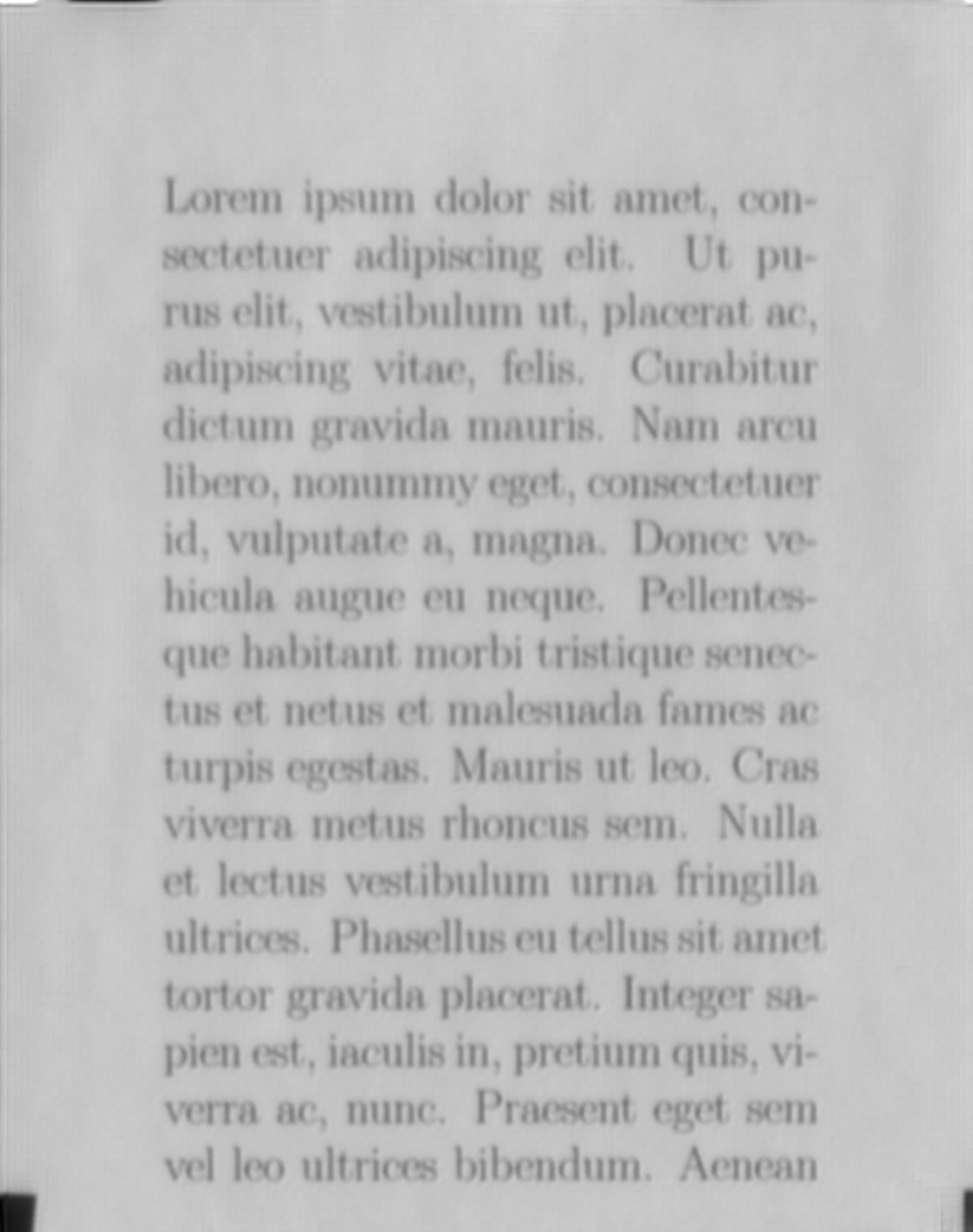} 
\end{subfigure}
\begin{subfigure}{0.32\textwidth}
\includegraphics[trim={8cm, 9.6cm, 8cm, 16cm}, clip, width=2.2in]{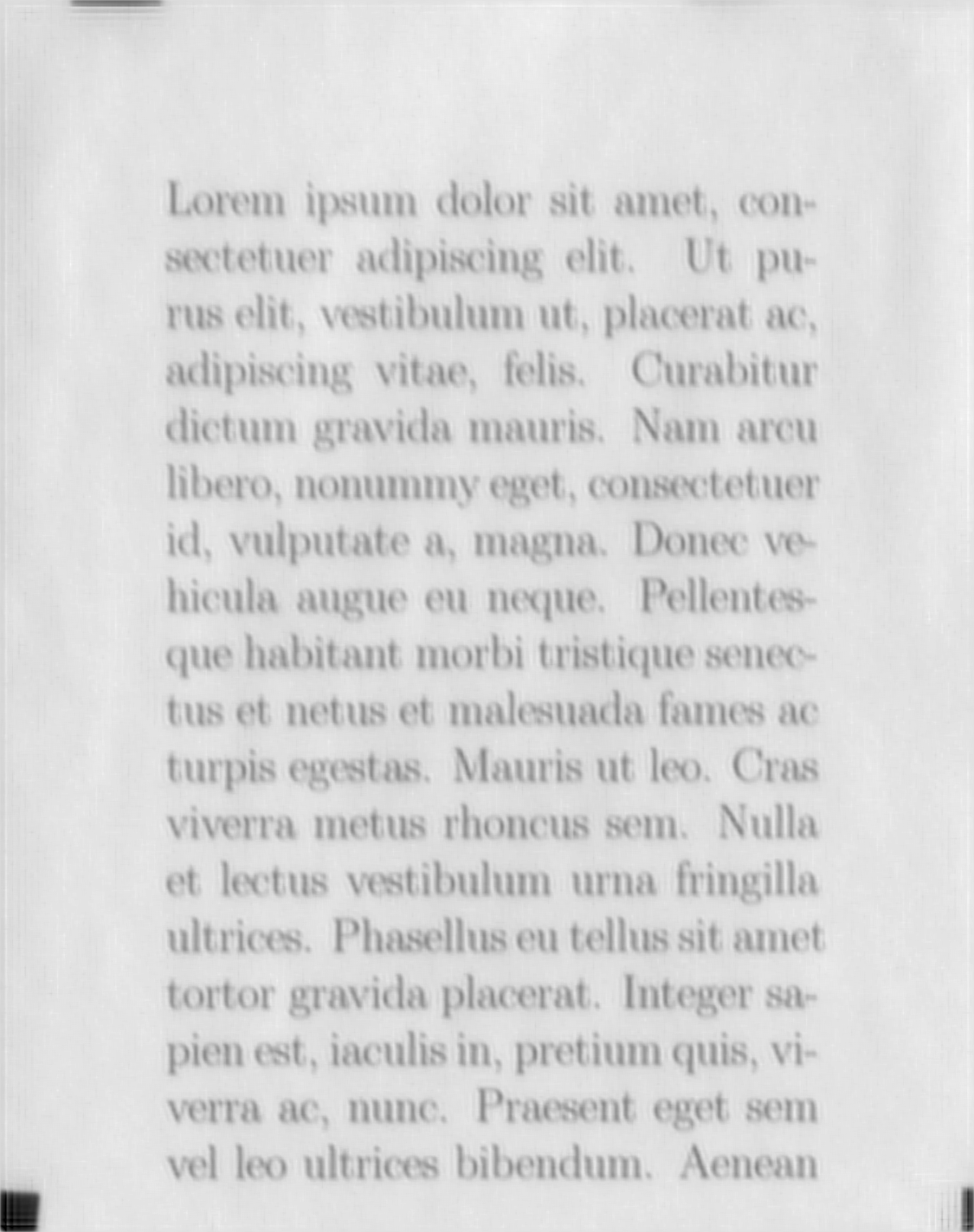} 
\end{subfigure}
\vspace{.2cm}

\begin{subfigure}{0.32\textwidth}
\includegraphics[trim={1cm, 1.2cm, 1cm, 2cm}, clip, width=2.2in]{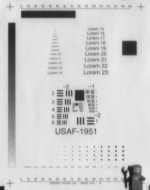} 
\end{subfigure}
\begin{subfigure}{0.32\textwidth}
\includegraphics[trim={8cm, 9.6cm, 8cm, 16cm}, clip, width=2.2in]{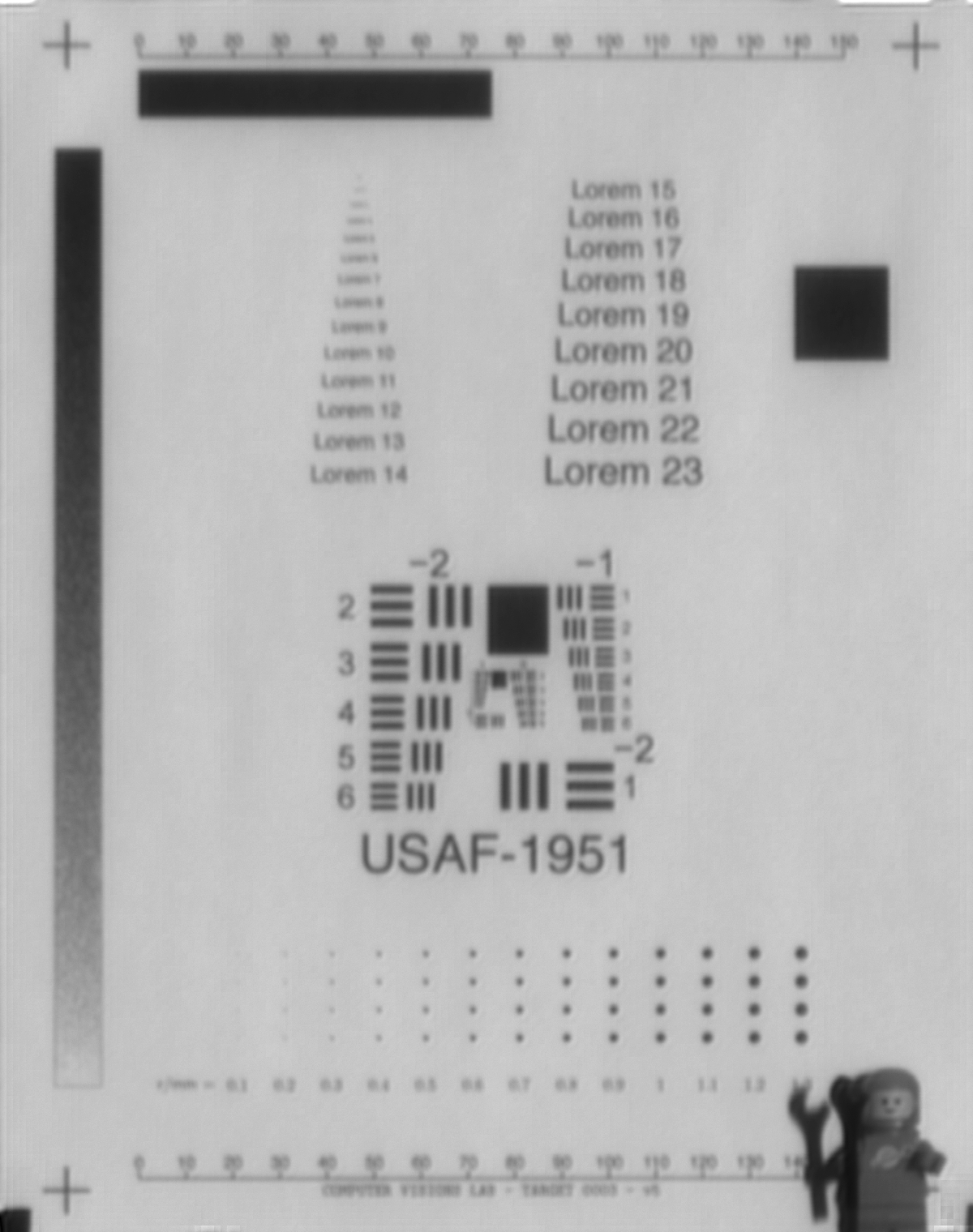} 
\end{subfigure}
\begin{subfigure}{0.32\textwidth}
\includegraphics[trim={8cm, 9.6cm, 8cm, 16cm}, clip, width=2.2in]{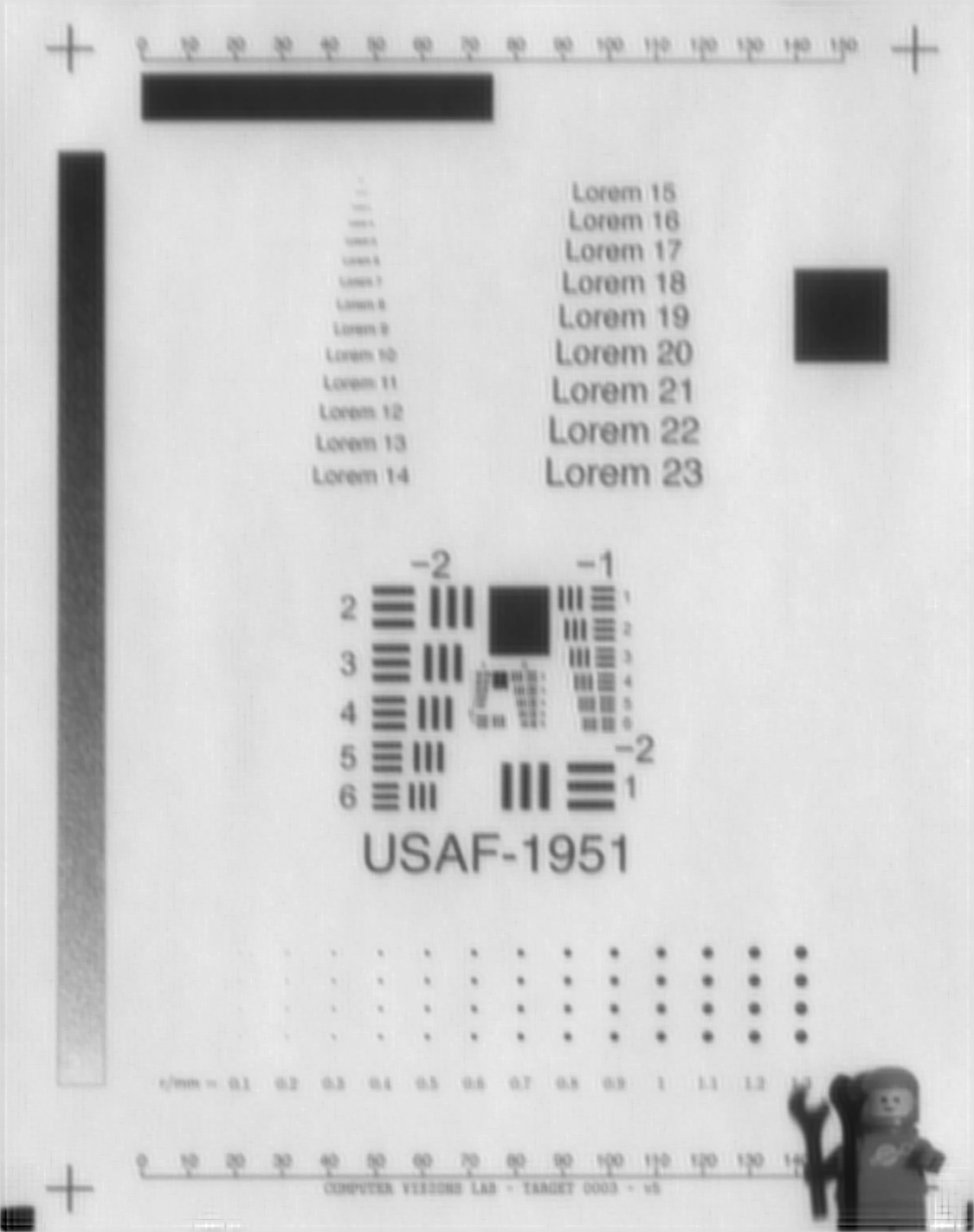} 
\end{subfigure}
\vspace{.2cm}

\begin{subfigure}{0.32\textwidth}
\includegraphics[trim={2.1cm, 0cm, 0.2cm, 3.45cm}, clip, width=2.2in]{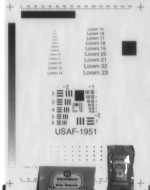} 
\caption{Static image (1 of 64)}
\end{subfigure}
\begin{subfigure}{0.32\textwidth}
\includegraphics[trim={16.8cm, 0cm, 1.6cm, 27.6cm}, clip, width=2.2in]{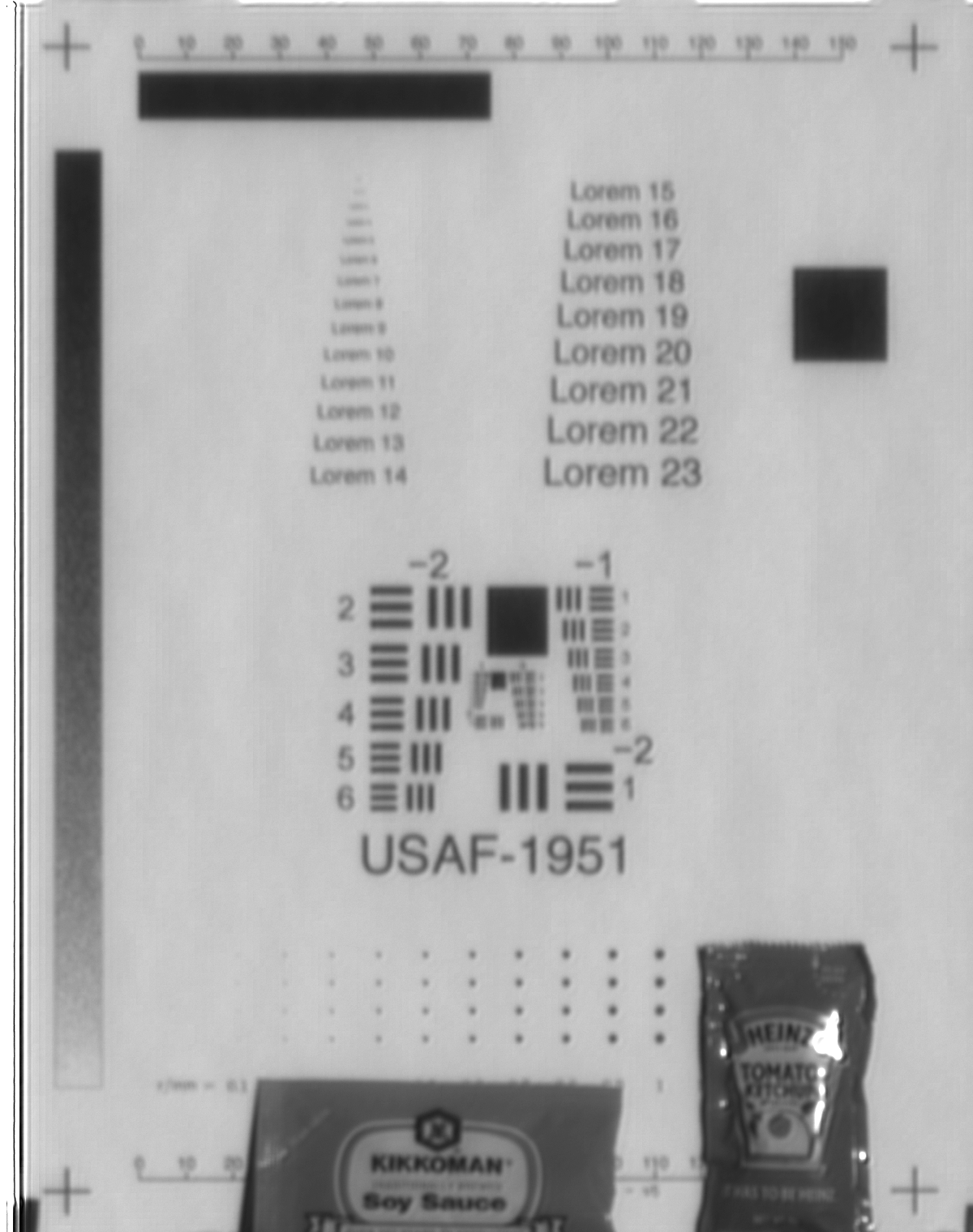} 
\caption{Wiener reconstruction}
\end{subfigure}
\begin{subfigure}{0.32\textwidth}
\includegraphics[trim={16.8cm, 0cm, 1.6cm, 27.6cm}, clip, width=2.2in]{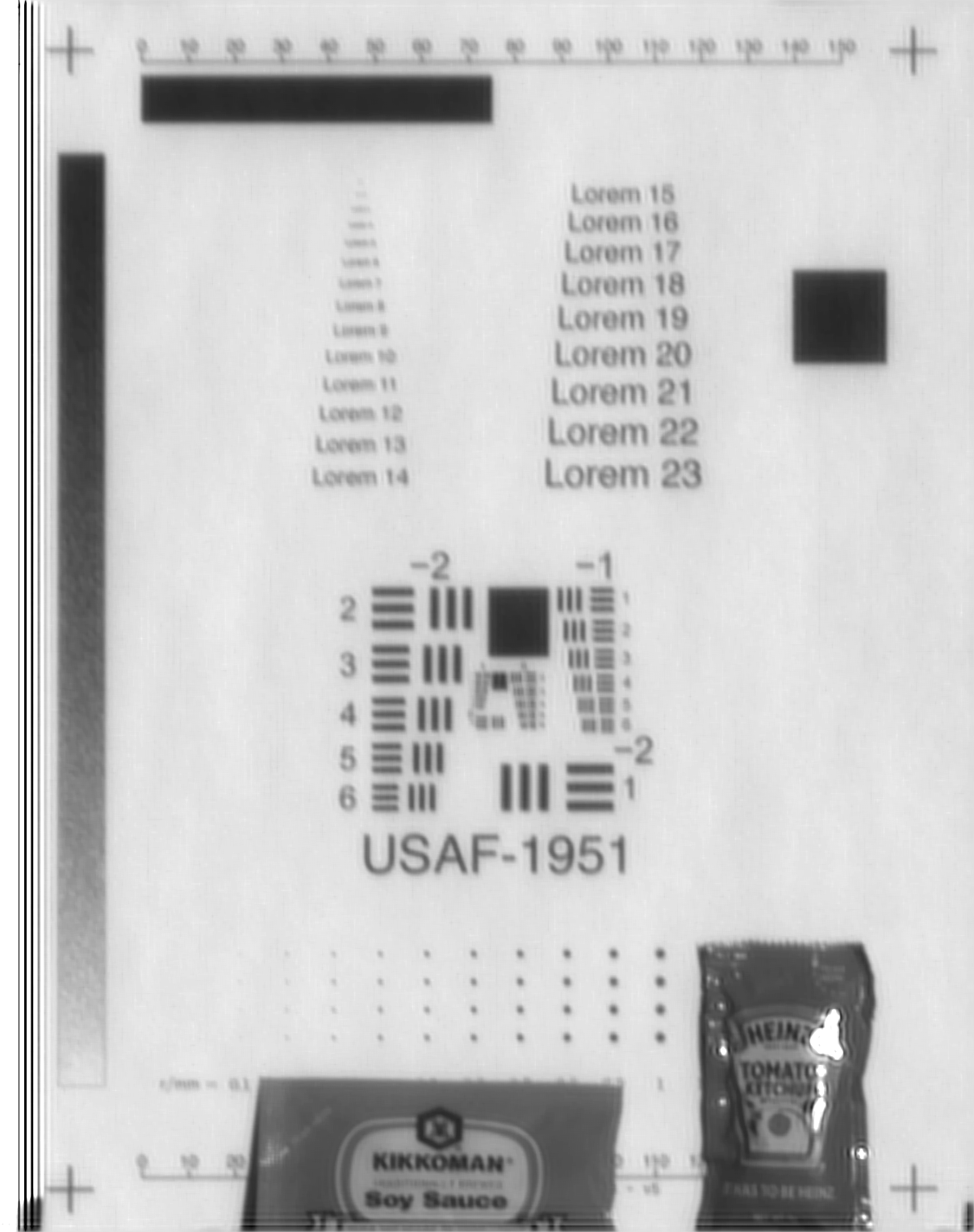} 
\caption{TV reconstruction}
\end{subfigure}
\hfill

\caption{Super-resolution using an 8x8 grid of static images.  We show
  a small crop of various targets.}
\label{fig:grid}
\end{figure*}

Figure~\ref{fig:grid} shows several results, each a small ROI of the
target area.  We note that in each example we can see high-resolution
details appear in the super-resolution images, such as the punctuation and to
some extent serifs in the text, elements in USAF group -1 in the resolution target, and the serrated top edge of the ketchup packet.

\begin{figure*}
\centering
\begin{subfigure}{0.32\textwidth}
\includegraphics[width=2.2in]{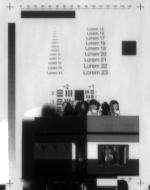} 
\end{subfigure}
\begin{subfigure}{0.32\textwidth}
\includegraphics[width=2.2in]{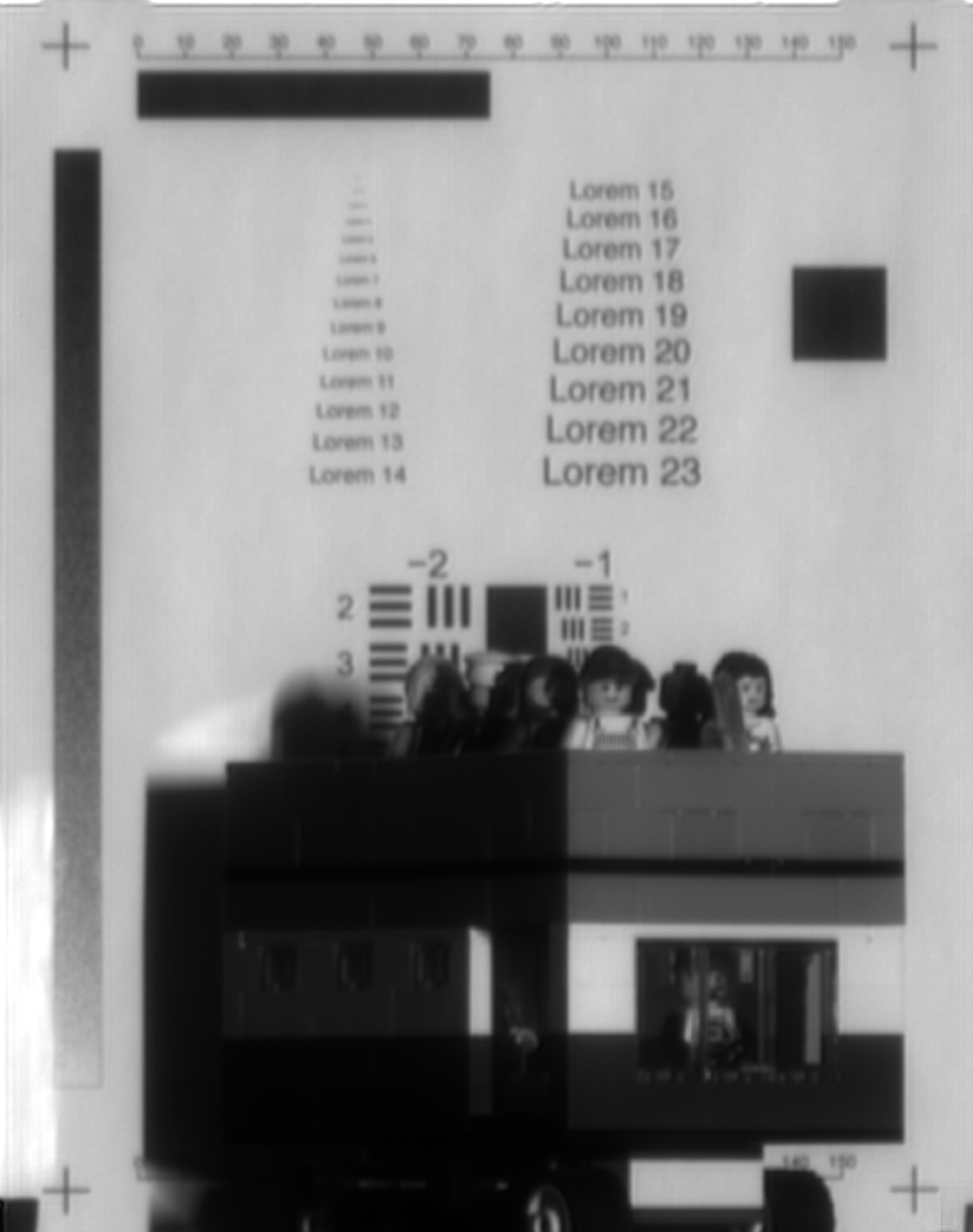} 
\end{subfigure}
\begin{subfigure}{0.32\textwidth}
\includegraphics[width=2.2in]{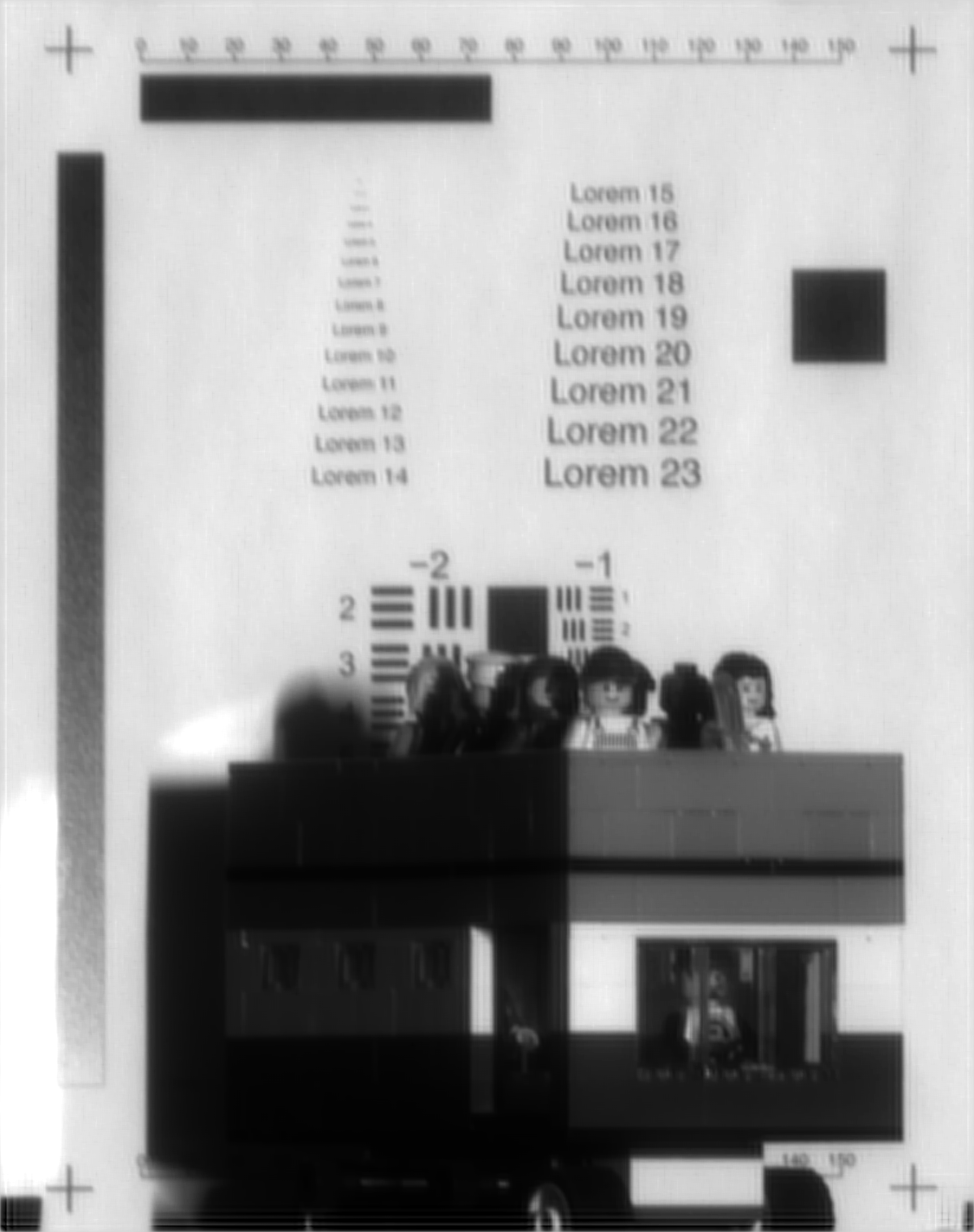} 
\end{subfigure}
\vspace{.1cm}

\begin{subfigure}{0.32\textwidth}
\includegraphics[trim={2cm, 1.6cm, 0.7cm, 2.3cm}, clip, width=2.2in]{figures/Bus-grid/crop-gray_0_0.png} 
\caption{Static image (1 of 64)}
\end{subfigure}
\begin{subfigure}{0.32\textwidth}
\includegraphics[trim={16cm, 12.8cm, 5.6cm, 18.4cm}, clip, width=2.2in]{figures/Bus-grid/reconstruction-fft-gray.png} 
\caption{Wiener reconstruction}
\end{subfigure}
\begin{subfigure}{0.32\textwidth}
\includegraphics[trim={16cm, 12.8cm, 5.6cm, 18.4cm}, clip, width=2.2in]{figures/Bus-grid/reconstruction-lops-gray.png} 
\caption{TV reconstruction}
\end{subfigure}
\hfill

\caption{Super-resolution using an 8x8 grid of static images.  The
  bottom row shows a crop of the front section of the bus. 
  Using a moving camera, instead
  of a moving sensor, requires the camera motion to be proportional to
  the distance from the object.  Nonetheless, we can image
  three-dimensional objects that are relatively close to a target
  plane.  In this case, a constant magnification model is still
  accurate.}
\label{fig:bus}
\end{figure*}

Moving the camera instead of the sensor can lead to motion parallax.  Figure~\ref{fig:bus} shows that a constant magnification model can 
still be used for
three-dimensional objects that protrude out of the target plane. The
front of the LEGO(R) bus is approximately 15 cm away from the
calibrated target plane and high-resolution details on the minifigures
are still resolved.  Even with twice as much depth variation, the parallax
effect would be small due to the large distance between the camera and the object.

Note that in Figure~\ref{fig:bus}, the striped overalls of a LEGO minifigure
appear uniform in a static image (a) but have visible stripes in the 
super-resolution image (c).  The stripes can be seen accurately as lines with
widths 3-4x smaller than the original pixel size, indicating a 3-4x improvement.

\begin{figure*}
\centering
\begin{subfigure}{0.32\textwidth}
\includegraphics[trim={16cm, 12.8cm, 5.6cm, 18.4cm}, clip, width=2.2in]{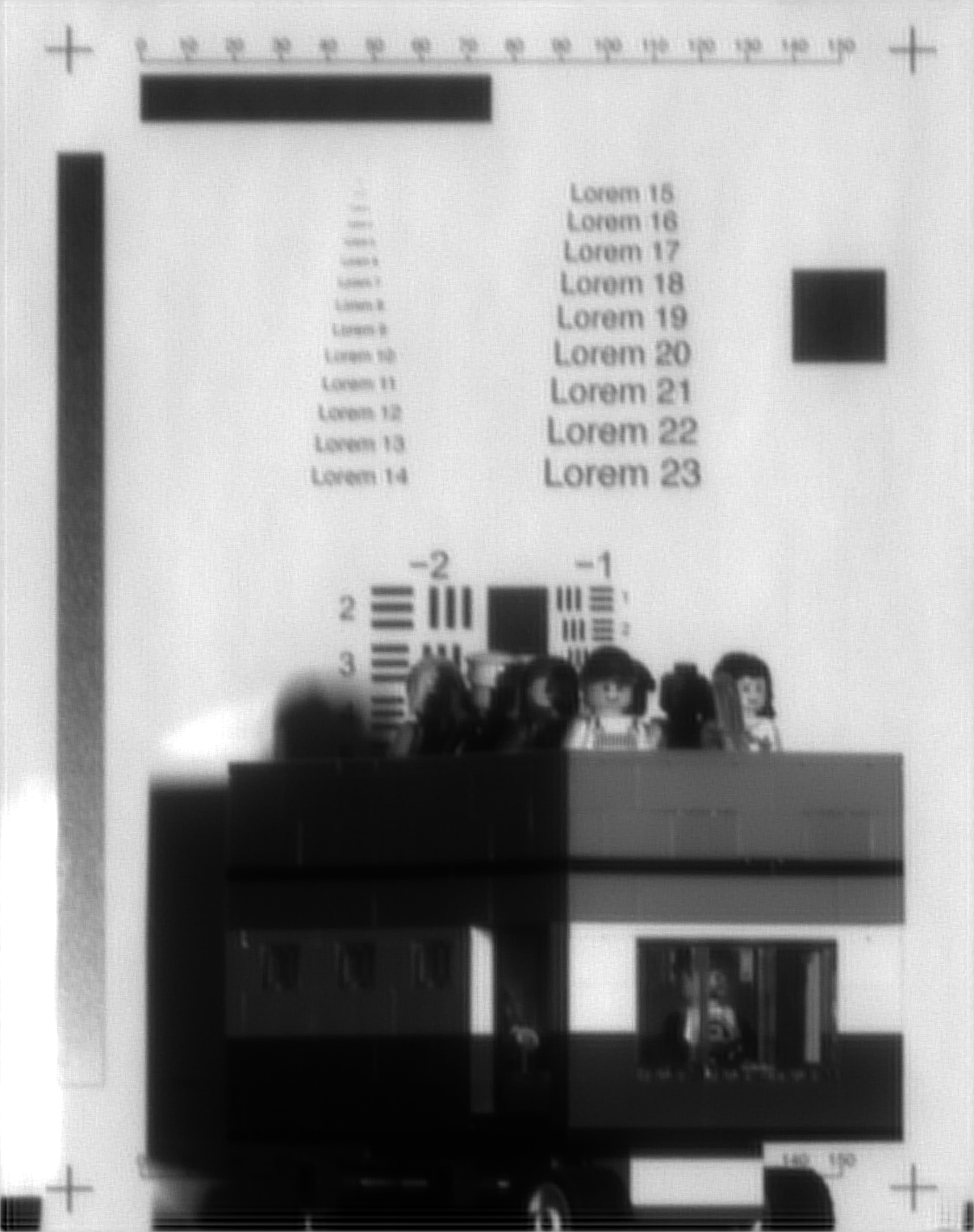} 
\caption{1 frame per position}
\end{subfigure}
\begin{subfigure}{0.32\textwidth}
\includegraphics[trim={16cm, 12.8cm, 5.6cm, 18.4cm}, clip, width=2.2in]{figures/Bus-grid/reconstruction-lops-gray.png} 
\caption{10 frames per position}
\end{subfigure}
\hfill

\caption{By averaging 10 frames per position, we reduce the effect of measurement noise (TV reconstructions).}
\label{fig:noise}
\end{figure*}

For the examples above, we averaged multiple frames taken from each camera position
to minimize the effect of measurement noise and to determine an upper limit on the
reconstruction quality obtained using a physical system.  
Figure~\ref{fig:noise} compares a reconstruction obtained using a single frame
from each position to a reconstruction obtained using ten frames from each
position.  While the reconstruction using a single frame from each position 
has more noise, it still recovers high-resolution features.

\subsection{Scan Motion}
\label{sec:scan}

To acquire images with motion, the stage was moved at constant
horizontal velocity over a “long” travel range corresponding to a
length of many pixels. As the stage moved, five images were captured
with a short random delay set before the start of each
acquisition. Once the motion across the row was completed, the height
of the stage was increased by sub-pixel steps (same as in grid
exposures) and the motion was repeated. The total number of heights
corresponds to the selected factor $f$.

Since our stage is slow, we used a relatively long exposure to capture
images with motion blur.  The exposure was determined by how long it
takes for the image on the sensor to move by $\Delta$ (the sensor pixel size).  The maximum velocity of the stage is only 2.5 mm/s and
therefore, for these experiments we used an exposure time of about
five hundred milliseconds.

Due to the fact that the stage has no position readout capabilities
and the camera cannot be synchronized with it, the initial camera
position for each acquisition must be estimated.  The exposure time, 
vertical position, and velocity of the stage during 
each capture are known.
The printed targets
include a black rectangle, which provides a step edge that is used for
the horizontal position estimation; similar to the blurred point source shown in
Figure 5, the initial position of a sharp 
edge can be localized to sub-pixel location from a moving camera. 

\begin{figure*}
\centering
\begin{subfigure}{0.35\textwidth}
\includegraphics[trim={1cm, 1.2cm, 1cm, 2cm}, clip, height=2.2in]{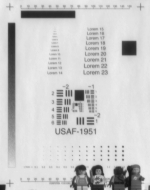} 
\end{subfigure}
\begin{subfigure}{0.35\textwidth}
\includegraphics[trim={4cm, 4.8cm, 4cm, 8cm}, clip, height=2.2in]{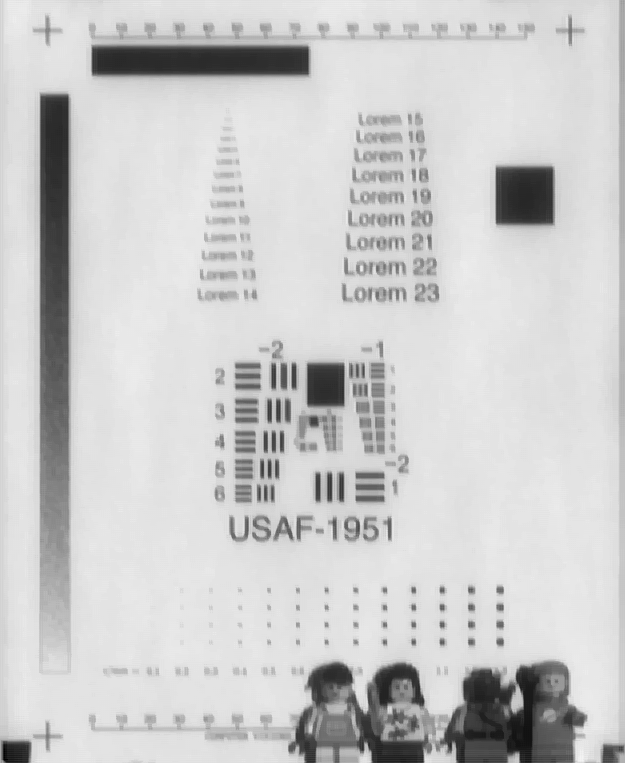} 
\end{subfigure}
\vspace{.2cm}

\begin{subfigure}{0.35\textwidth}
\includegraphics[trim={1cm, 0cm, 1cm, 3.2cm}, clip, height=2.2in]{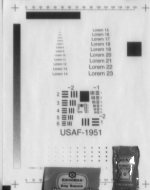} 
\caption{Static image}
\end{subfigure}
\begin{subfigure}{0.35\textwidth}
\includegraphics[trim={4cm, 0cm, 4cm, 12.8cm}, clip, height=2.2in]{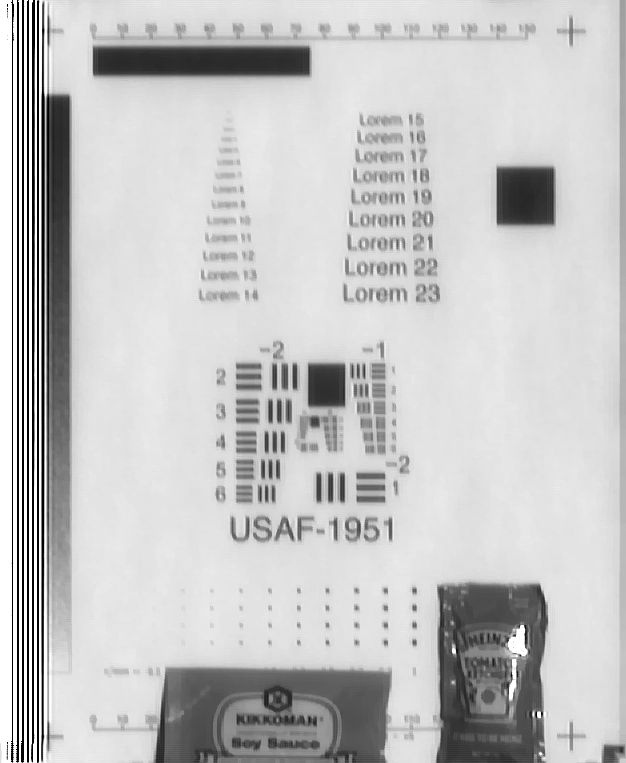} 
\caption{TV reconstruction}
\end{subfigure}

\caption{Super-resolution using 4 horizontal scans with constant
  velocity and 4 exposures per scan.  (a) shows an ROI from a static
  image.  (b) shows the super-resolution result using 16 images taken
  with constant velocity motion.}
\label{fig:scan}
\end{figure*}

Figure~\ref{fig:scan} illustrates some results of the scanning motion
procedure.  These examples demonstrate the use of a moving camera to
increase resolution beyond what is visible in a static image, despite
the induced motion blur.

\section{Conclusion and Future Work}

We described several results related to the limits of
super-resolution with a moving sensor/camera.  

Our contributions include:

1) A new analysis and demonstration of super-resolution and
deconvolution with a box using sparse image priors.

2) The idea and demonstration that motion blur
can be beneficial for super-resolution.

\begin{figure*}
\centering
\begin{subfigure}{0.3\textwidth}
\includegraphics[width=2.5in]{./figures/discrete-periodic-box-ft4.png} 
\caption{$|\fft(B_4)|$ (in dB)}
\end{subfigure}
\begin{subfigure}{0.3\textwidth}
\includegraphics[width=2.5in]{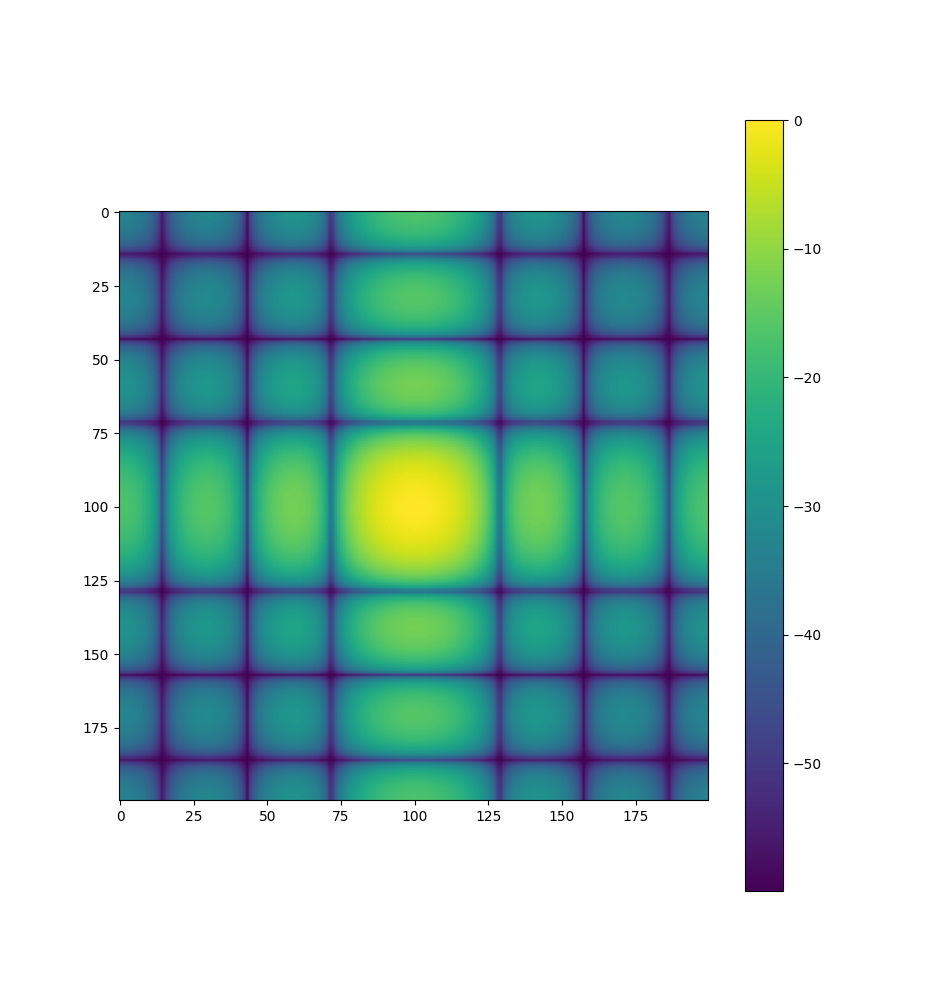} 
\caption{$|\fft(B_7)|$ (in dB)}
\end{subfigure}
\begin{subfigure}{0.3\textwidth}
\includegraphics[width=2.5in]{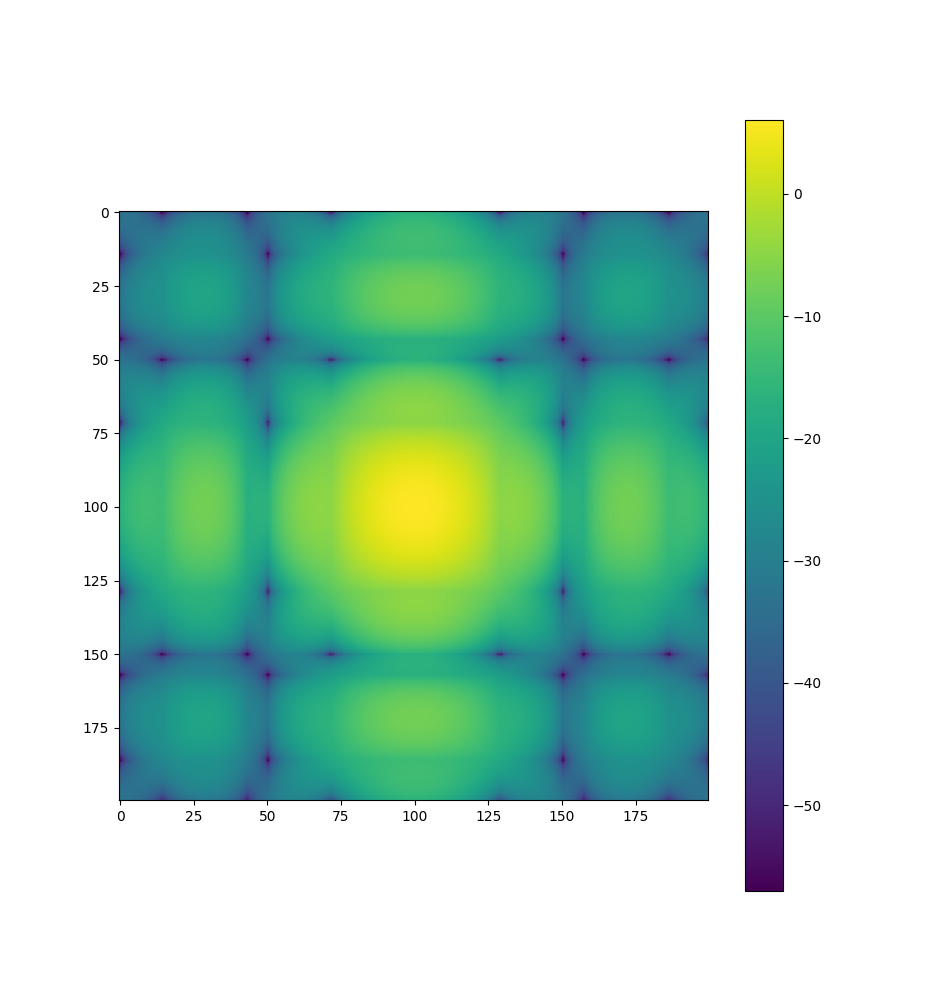} 
\caption{$|\fft(B_4)|+|\fft(B_7)|$ (in dB)}
\end{subfigure}
\caption{Bode plots of a two-dimensional box filter of width 4 (a),
  width 7 (b), and the sum of the two (c).  The last plot illustrates
  how using pixels of two different sizes should eliminate most of the
  ambiguity for super-resolution.}
\label{fig:box-ft}
\end{figure*}

We have shown that by using a grid of static images, the ambiguity in 
super-resolution can be
restricted to relatively few Fourier coefficients.  It is interesting to note 
that by using multiple magnifications, or
a sensor with multiple pixel sizes, one could potentially remove all
of the ambiguity in the super-resolution problem.
Figure~\ref{fig:box-ft} illustrates the Fourier transform of a
discrete periodic box of width 4 and similarly of a box of width 7.
We see that the zeros of the two Fourier transforms occur in different
places; if we were to record the convolution of a high-resolution
image with boxes of two sizes, we would eliminate essentially all
ambiguity in the super-resolution problem.  

The idea that motion blur can be used to improve resolution is somewhat
counterintuitive and leads to the notion that to take
a sharper picture, one should shake or vibrate the camera.  In
synthetic experiments, we have shown that simply vibrating the image
sensor during capture can lead to higher resolution images.  In this
case reconstruction involves solving a deconvolution problem while
increasing resolution at the same time.

\bibliographystyle{IEEEtran}
\bibliography{blurry}

\begin{thebibliography}{10}
\providecommand{\url}[1]{#1}
\csname url@samestyle\endcsname
\providecommand{\newblock}{\relax}
\providecommand{\bibinfo}[2]{#2}
\providecommand{\BIBentrySTDinterwordspacing}{\spaceskip=0pt\relax}
\providecommand{\BIBentryALTinterwordstretchfactor}{4}
\providecommand{\BIBentryALTinterwordspacing}{\spaceskip=\fontdimen2\font plus
\BIBentryALTinterwordstretchfactor\fontdimen3\font minus
  \fontdimen4\font\relax}
\providecommand{\BIBforeignlanguage}[2]{{%
\expandafter\ifx\csname l@#1\endcsname\relax
\typeout{** WARNING: IEEEtran.bst: No hyphenation pattern has been}%
\typeout{** loaded for the language `#1'. Using the pattern for}%
\typeout{** the default language instead.}%
\else
\language=\csname l@#1\endcsname
\fi
#2}}
\providecommand{\BIBdecl}{\relax}
\BIBdecl

\bibitem{baker_limits_2002}
S.~Baker and T.~Kanade, ``Limits on super-resolution and how to break them,''
  \emph{IEEE transactions on pattern analysis and machine intelligence},
  vol.~24, no.~9, pp. 1167--1183, 2002.

\bibitem{lin_fundamental_2004}
Z.~Lin and H.-Y. Shum, ``Fundamental limits of reconstruction-based
  superresolution algorithms under local translation,'' \emph{IEEE transactions
  on pattern analysis and machine intelligence}, vol.~26, no.~1, pp. 83--97,
  2004.

\bibitem{ben-ezra_jitter_2004}
M.~Ben-Ezra, A.~Zomet, and S.~K. Nayar, ``Jitter camera: {High} resolution
  video from a low resolution detector,'' in \emph{Proceedings of the 2004 IEEE
  Computer Society Conference on Computer Vision and Pattern Recognition},
  vol.~2.

\bibitem{ben-ezra_high_2010}
M.~Ben-Ezra, ``High resolution large format tile-scan camera: {Design},
  calibration, and extended depth of field,'' in \emph{2010 {International}
  {Conference} on {Computational} {Photography}}.\hskip 1em plus 0.5em minus
  0.4em\relax IEEE, pp. 1--8.

\bibitem{irani-peleg-super}
M.~Irani and S.~Peleg, ``Super resolution from image sequences,'' in
  \emph{Proceedings. 10th International Conference on Pattern Recognition},
  vol.~2.\hskip 1em plus 0.5em minus 0.4em\relax IEEE, 1990, pp. 115--120.

\bibitem{zisserman_capel_text}
D.~Capel and A.~Zisserman, ``Super-resolution enhancement of text image
  sequences,'' in \emph{Proceedings: 15th International Conference on Pattern
  Recognition.}\hskip 1em plus 0.5em minus 0.4em\relax IEEE, 2000.

\bibitem{candes_robust_2006}
E.~J. Candès, J.~Romberg, and T.~Tao, ``Robust uncertainty principles: {Exact}
  signal reconstruction from highly incomplete frequency information,''
  \emph{IEEE Transactions on information theory}, vol.~52, no.~2, pp. 489--509,
  2006.

\bibitem{donoho1989-uncertainty}
D.~L. Donoho and P.~B. Stark, ``Uncertainty principles and signal recovery,''
  \emph{SIAM Journal on Applied Mathematics}, vol.~49, no.~3, pp. 906--931,
  1989.

\bibitem{romberg2009-compressive}
J.~Romberg, ``Compressive sensing by random convolution,'' \emph{SIAM Journal
  on Imaging Sciences}, vol.~2, no.~4, pp. 1098--1128, 2009.

\bibitem{tropp2006random-filters}
J.~A. Tropp, M.~B. Wakin, M.~F. Duarte, D.~Baron, and R.~G. Baraniuk, ``Random
  filters for compressive sampling and reconstruction,'' in \emph{2006 IEEE
  International Conference on Acoustics Speech and Signal Processing
  Proceedings}, vol.~3.

\bibitem{fergus2006-random-lens}
R.~Fergus, A.~Torralba, and W.~T. Freeman, ``Random lens imaging,''
  Massachusetts Institute of Technology Computer Science and Artificial
  Intelligence Laboratory, Tech. Rep., 2006.

\bibitem{antipa2017-diffusercam}
N.~Antipa, G.~Kuo, R.~Heckel, B.~Mildenhall, E.~Bostan, R.~Ng, and L.~Waller,
  ``Diffuser{C}am: lensless single-exposure 3d imaging,'' \emph{Optica},
  vol.~5, no.~1, pp. 1--9, 2017.

\bibitem{duarte2008-single-pixel}
M.~F. Duarte, M.~A. Davenport, D.~Takhar, J.~N. Laska, T.~Sun, K.~F. Kelly, and
  R.~G. Baraniuk, ``Single-pixel imaging via compressive sampling,'' \emph{IEEE
  signal processing magazine}, vol.~25, no.~2, pp. 83--91, 2008.

\bibitem{levin_motion-invariant_2008}
A.~Levin, P.~Sand, T.~S. Cho, F.~Durand, and W.~T. Freeman, ``Motion-invariant
  photography,'' \emph{ACM Transactions on Graphics}, vol.~27, no.~3, pp. 1--9,
  2008.

\bibitem{raskar2006-coded-flutter}
R.~Raskar, A.~Agrawal, and J.~Tumblin, ``Coded exposure photography: motion
  deblurring using fluttered shutter,'' \emph{ACM Transactions on Graphics},
  vol.~25, no.~3, pp. 795--804, 2006.

\bibitem{tendero_flutter_2013}
Y.~Tendero, J.-M. Morel, and B.~Rougé, ``The flutter shutter paradox,''
  \emph{SIAM Journal on Imaging Sciences}, vol.~6, no.~2, pp. 813--847, 2013.

\bibitem{hayes-sun2012-super}
L.~Sun and J.~Hays, ``Super-resolution from internet-scale scene matching,'' in
  \emph{2012 IEEE International conference on computational photography}, pp.
  1--12.

\bibitem{glasner_super-resolution_2009}
D.~Glasner, S.~Bagon, and M.~Irani, ``Super-resolution from a single image,''
  in \emph{12th international conference on computer vision}.\hskip 1em plus
  0.5em minus 0.4em\relax IEEE, 2009, pp. 349--356.

\bibitem{dong2015-deep}
C.~Dong, C.~C. Loy, K.~He, and X.~Tang, ``Image super-resolution using deep
  convolutional networks,'' \emph{IEEE transactions on pattern analysis and
  machine intelligence}, vol.~38, no.~2, pp. 295--307, 2015.

\bibitem{kamilov2023plug-play-priors}
U.~S. Kamilov, C.~A. Bouman, G.~T. Buzzard, and B.~Wohlberg, ``Plug-and-play
  methods for integrating physical and learned models in computational imaging:
  Theory, algorithms, and applications,'' \emph{IEEE Signal Processing
  Magazine}, vol.~40, no.~1, pp. 85--97, 2023.

\bibitem{xie_neural_2022}
Y.~Xie, T.~Takikawa, S.~Saito, O.~Litany, S.~Yan, N.~Khan, F.~Tombari,
  J.~Tompkin, V.~Sitzmann, and S.~Sridhar, ``Neural fields in visual computing
  and beyond,'' in \emph{Computer {Graphics} {Forum}}, vol.~41.\hskip 1em plus
  0.5em minus 0.4em\relax Wiley Online Library, 2022, pp. 641--676, issue: 2.

\bibitem{wronski2019-handheld}
B.~Wronski, I.~Garcia-Dorado, M.~Ernst, D.~Kelly, M.~Krainin, C.-K. Liang,
  M.~Levoy, and P.~Milanfar, ``Handheld multi-frame super-resolution,''
  \emph{ACM Transactions on Graphics}, vol.~38, no.~4, pp. 1--18, 2019.

\bibitem{felzenszwalb_2024}
\BIBentryALTinterwordspacing
P.~Felzenszwalb, ``Deconvolution with a box,'' 2024. [Online]. Available:
  \url{https://arxiv.org/abs/2407.11685}
\BIBentrySTDinterwordspacing

\bibitem{chambolle2010introduction}
A.~Chambolle, V.~Caselles, D.~Cremers, M.~Novaga, and T.~Pock, ``An
  introduction to total variation for image analysis,'' in \emph{Theoretical
  foundations and numerical methods for sparse recovery}, M.~Fornasier, Ed.,
  2010, pp. 263--340.

\bibitem{schiebinger2018superresolution-separation}
G.~Schiebinger, E.~Robeva, and B.~Recht, ``Superresolution without
  separation,'' \emph{Information and Inference: A Journal of the IMA}, vol.~7,
  no.~1, pp. 1--30, 2018.

\end{thebibliography}

\end{document}